\documentclass[nohyperref]{article}

\usepackage{microtype}
\usepackage{graphicx}
\usepackage{subfigure}
\usepackage{booktabs} 

\usepackage{wrapfig}

\usepackage{hyperref}

\usepackage[accepted]{icml2022}

\usepackage{amsmath}
\usepackage{amssymb}
\usepackage{mathtools}
\usepackage{amsthm}

\usepackage[capitalize,noabbrev]{cleveref}

\theoremstyle{plain}
\newtheorem{theorem}{Theorem}[section]

\newtheorem{lemma}[theorem]{Lemma}
\newtheorem{corollary}[theorem]{Corollary}
\theoremstyle{definition}

\theoremstyle{remark}

\DeclareMathOperator*{\sgn}{sgn}
\DeclareMathOperator*{\Top-k}{Top-k}
\DeclareMathOperator{\rank}{rank}
\DeclareMathOperator\erf{erf}

\usepackage[textsize=tiny]{todonotes}

\icmltitlerunning{MESH: Memory Scaffold with Heteroassociation}

\begin{document}

\setlength{\abovedisplayskip}{4pt}
\setlength{\belowdisplayskip}{4pt}


\twocolumn[
\icmltitle{Content Addressable Memory Without Catastrophic Forgetting by Heteroassociation with a Fixed Scaffold}





%



\icmlsetsymbol{equal}{*}

\begin{icmlauthorlist}
\icmlauthor{Sugandha Sharma}{mit}
\icmlauthor{Sarthak Chandra}{equal,mit}
\icmlauthor{Ila R. Fiete}{equal,mit}
\end{icmlauthorlist}

\icmlaffiliation{mit}{Department of Brain and Cognitive Sciences, McGovern Institute for Brain Research, \& Integrative Computational Neuroscience Center (ICoN), Massachusetts Institute of Technology, Cambridge, USA}

\icmlcorrespondingauthor{Sugandha Sharma}{susharma@mit.edu}

\icmlkeywords{Machine Learning, ICML}

\vskip 0.3in
]



\printAffiliationsAndNotice{\icmlEqualContribution}  

\begin{abstract}
\vspace*{-.2em}
Content-addressable memory (CAM) networks, so-called because stored items can be recalled by partial or corrupted versions of the items, exhibit near-perfect recall of a small number of information-dense patterns below capacity and a `memory cliff' beyond, such that inserting a single additional pattern results in catastrophic loss of all stored patterns. We propose a novel CAM architecture, Memory Scaffold with Heteroassociation (MESH), that factorizes the problems of internal attractor dynamics and association with external content to generate a CAM continuum without a memory cliff: Small numbers of patterns are stored with complete information recovery matching standard CAMs, while inserting more patterns still results in partial recall of every pattern, with a graceful trade-off between pattern number and pattern richness. Motivated by the architecture of the Entorhinal-Hippocampal memory circuit in the brain, MESH is a tripartite architecture with pairwise interactions that uses a predetermined set of internally stabilized states together with heteroassociation between the internal states and arbitrary external patterns. We show analytically and experimentally that for any number of stored patterns, MESH nearly saturates the total information bound (given by the number of synapses) for CAM networks, outperforming all existing CAM models. 
\end{abstract}
\vspace{-6mm}
\section{Introduction}
\label{sec:intro}


\begin{figure*}[h]
\centering
\includegraphics[width=0.85\textwidth]{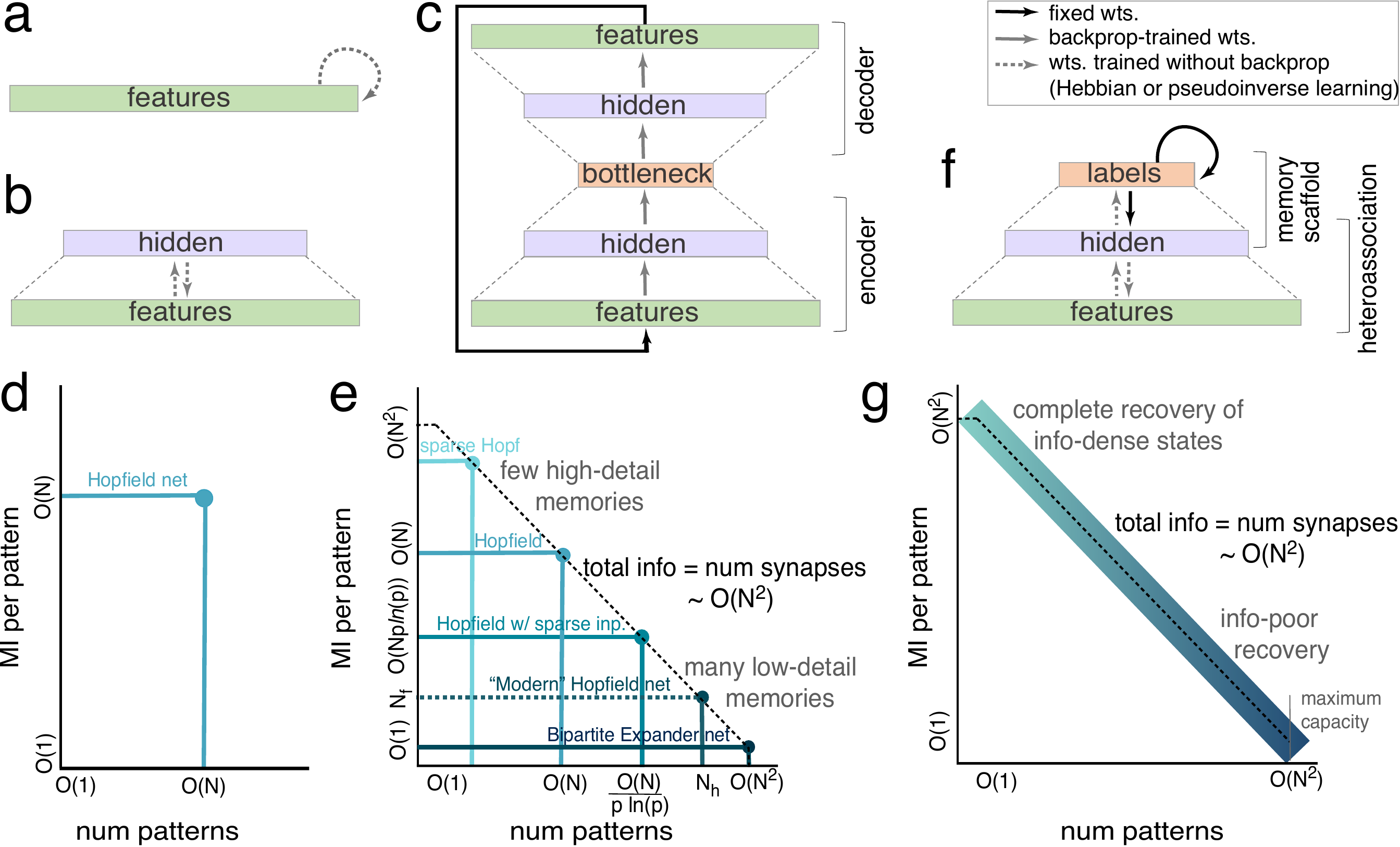}
\caption{\textbf{Content-addressable memory (CAM) architectures, the memory cliff, and the CAM continuum.} (a-c) Existing CAM  architectures. (a) A fully recurrently connected Hopfield CAM network \cite{hopfield1982neural, hopfield1984neurons}. (b) Bipartite CAM networks: Bipartite Expander Hopfield Network \cite{chaudhuri2019bipartite}, Modern Hopfield Network \cite{krotov2020large}. (c) Overparameterized tail-biting autoencoder as a CAM~\cite{radhakrishnan2020overparameterized}. (d) Schematic of the memory cliff exhibited by Hopfield networks: addition of a pattern beyond the critical capacity leads to catastrophic forgetting (loss) of all patterns. (e) Theoretical upper-bound envelope for storage of binary patterns in CAM networks with $N^2$ synapses (black dashed line); depending on the architecture, the number of nodes varies across networks. Existing networks each exhibit a memory cliff, and approach the envelope at only one point (i.e., at a specific number of stored patterns; Modern Hopfield Networks exist only at this point since the number of hidden units must \emph{exactly} equal the number of memories as we denote by the dashed line). (f) Tripartite architecture of our proposed model, MESH. (g) The desired CAM continuum: a single network with information storage near the theoretical bound envelope regardless of the number of stored patterns.}
\label{fig:schematics}
\end{figure*} 

Content-addressable memory (CAM) networks are compelling models of long-term human memory: Humans are experts at recognizing situations or items they have encountered before, and often fill in the details from partial or noisy information. Similarly, when presented by a partial or corrupted version of a previously memorized input, a recurrently iterated (autoassociative) CAM can reconstruct the learned pattern by flow to fixed points. For example, the Hopfield network \cite{hopfield1982neural, hopfield1984neurons} encodes the memorized states as fixed points of its dynamics. Because the state remains at a fixed point once it reaches there, such CAM networks can thus also function as short-term memory networks for the acquired long-term memories.

Several recurrent network architectures support CAM dynamics, including the Hopfield network \cite{hopfield1982neural, hopfield1984neurons} (Fig.~\ref{fig:schematics}a), several variants of the Hopfield network \cite{personnaz1985information, tsodyks1988enhanced, krotov2020large} (Fig.~\ref{fig:schematics}a,b), and overparameterized autoencoders \cite{radhakrishnan2020overparameterized} (Fig.~\ref{fig:schematics}c). However, all of them exhibit a memory cliff, beyond which adding a single pattern leads to catastrophic loss of all patterns  (Fig.~\ref{fig:schematics}d).  The total information content of CAM networks is bounded theoretically by $\mathcal{O}(N^2)$, the number of synapses in the network~\cite{abu1989information, gardner1988space}, Fig.~\ref{fig:schematics}e, defining a total information budget to be split between the number of stored patterns and information per pattern. However, most CAM networks approach that bound only when storing a fixed, specific number of pattterns (Fig.~\ref{fig:schematics}e): Different CAM networks (defined by their inputs, architecture, or weight and activity update rules) touch this total information envelope at different points, with some storing a small number of maximally detailed memory states, others storing a larger number of less-detailed states. None of these models have the flexibility to span the memory envelope such that the information recalled per pattern is continuously traded off for increasing numbers of stored patterns in an online way, while preserving a constant total information that remains close to the information envelope. 

In this paper we propose a novel and biologically motivated memory architecture, Memory Scaffold with Heteroassociation (MESH), that generates a CAM continuum (see Fig.~\ref{fig:schematics}f for a schematic of the network architecture). 
MESH factorizes the problem of associative memory into two separate pieces: a part that does memory through a pre-defined ``\emph{memory scaffold}'', and a part that does association through a ``\emph{heteroassociative}'' step.
Inspired by the Entorhinal-Hippocampal memory system in mammalian brains, MESH contains a bipartite attractor network with random weights that stabilizes a large dictionary of well-separated and pre-defined fixed points that serve as the memory scaffold \cite{10.7554/eLife.62702, Mulders2021.11.20.469406}.
Arbitrary dense patterns are then stored by heteroassociatively linking them to the pre-defined scaffold states. 

MESH can be viewed as a reservoir network for memory, in the sense that predefined recurrently determined fixed points are associated with arbitrary inputs to perform storage, similar to the way in which standard reservoir networks associate a predefined recurrently determined dynamical system to dynamical trajectories in training data~\cite{jaeger2001echo,lukovsevivcius2009reservoir}.

This novel combination  results in a neural network with a CAM continuum (CAMC) that approaches the theoretical upper-bound on information storage \cite{abu1989information, gardner1988space} regardless of the number of stored patterns (Fig.~\ref{fig:schematics}g). Storage of information-dense patterns up to a critical capacity results in complete recovery of all patterns and storage of larger numbers of patterns results in partial reconstruction of the corresponding stored pattern. Partial reconstruction continues up to an exponentially large number of patterns as a function of total number of neurons in the network, ending in correct recognition of exponentially many stored patterns. To our knowledge, this is the first model of a CAM that automatically trades off pattern number and pattern richness. It predicts that biological memory systems may exploit pre-existing scaffolds to acquire new memories, potentially consistent with the observed preplay of hippocampal sequences before they are used for representing new environments \cite{dragoi2011preplay}. 


In the next section, we discuss existing CAM models and their dynamics. In Section~\ref{sec:MESH_CAMC} we provide our central results on the memory continuum exhibited by MESH. In Sections~\ref{sec:memory_scaffold} and~\ref{sec:arbitrary_patts}, we analyze how MESH works. In section~\ref{sec:applications} we extend MESH to the case of continuous neural activations, apply it to a realistic dataset, and show that it continues to exhibit the memory continuum even when storing continuous patterns.

\section{Existing CAM Models Lack a Memory Continuum}

\begin{figure*}[h]
\centering
\includegraphics[width=0.95\textwidth]{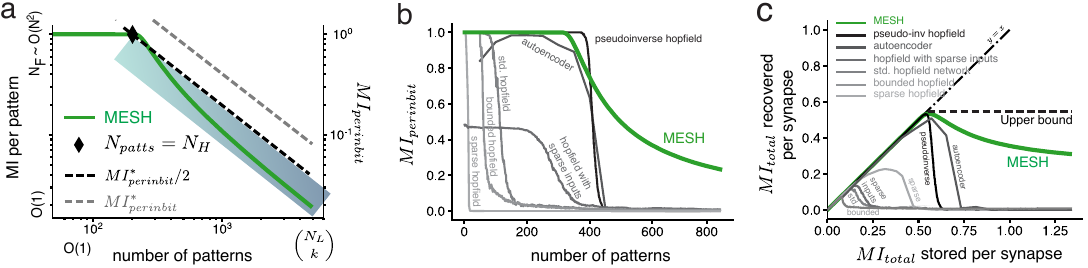}
\caption{\textbf{MESH exhibits a near-optimal CAM continuum.} (a) Mutual information per input bit between the stored and recovered patterns in MESH, as a function of the number of patterns stored in the network. MESH shows gradual degradation of mutual information upper-bounded by the theoretical upper bound, and can store upto $\binom{N_{L}}{k} \sim e^{dN_L}$ states without confusing the states and with non-zero level of detail, with a number of synapses and neurons that grows linearly with $\binom{N_{L}}{k}$. Here $N_L = 32$, $k=3$, $N_H = 200$, $N_F = 4960$. (b) Mutual information (per input bit) in existing networks that show catastrophic forgetting relative to MESH (see Fig.~\ref{fig:appendix_MIrelative}b for mutual information as a function of information stored per synapse). (c) Comparison of information per synapse across different networks relative to MESH. Given a network of fixed size, the total information in MESH is invariant to the number of stored patterns, while it decays to zero for other networks. In (b) and (c) all shown networks have $\approx 5\times 10^5$ synapses, with MESH layer sizes: $N_L = 18$, $N_H = 300$, $N_F = 816$, and $k = 3$ active bits in the label layer. See Fig.~\ref{fig:existing_nets} for the number of nodes in other networks.
MESH curves in (a), (b), (c) are averaged over 20 runs with different random initializations of the predefined connectivity, error bars are too small to be visible.
}
\label{fig:MESH_CAM_resutls}
\end{figure*}


Here we review  existing CAM architectures. Unless otherwise specified, we consider networks with $N$ neurons and dense binary activations (i.e., $N$-dimensional vectors with activations of 1 or -1 in each entry). 


Hopfield networks~\cite{hopfield1982neural} (Fig.~\ref{fig:schematics}a) can store up to $\approx0.14N$ random binary patterns. Beyond this capacity, the network demonstrates a memory cliff~\cite{nadal1986networks, crisanti1986saturation} (Fig.~\ref{fig:existing_nets}a). The recurrent weights in the Hopfield network may be set by a pseudoinverse learning rule~\cite{personnaz1985information}, where the network is guaranteed to store up to $N$ linearly independent patterns. However, storing more than $N/2$ patterns results in vanishing basins of attraction around each fixed point~\cite{personnaz1986collective, kanter1987associative} (Fig.~\ref{fig:existing_nets}b). Bounded synapse models~\cite{parisi1986memory, fusi2007limits, van2012soft} on the other hand, do not exhibit a memory cliff in the same sense as the classic Hopfield network, however, attempted storage of a large number of patterns results in complete loss of a large fraction of the stored patterns with only $\approx0.04N$ patterns correctly recalled (Fig.~\ref{fig:existing_nets}c).

Hopfield networks with sparse inputs store sparse $\{0,1\}$ binary patterns with a fraction $p$ of non-zero entries, instead of the usual dense $\{-1,1\}$ patterns~\cite{tsodyks1988enhanced}. They can alternatively store a larger number of sparse patterns, given by $(p|\ln(p)|)^{-1}N$, such that the product of number of patterns times information per pattern is constant. However, the tradeoff between pattern number and pattern information arises for differently structured sets of input patterns rather than from flexibility of the network -- every stored pattern is either fully recalled or all patterns are lost to a memory cliff at the pattern capacity (Fig.~\ref{fig:existing_nets}d). In other words, a single network presented with a single type (sparsity level) of data does not exhibit a tradeoff between pattern number and pattern information. Sparse Hopfield networks have sparse connectivity~\cite{dominguez2007information} but store dense $\{-1,1\}$ patterns. These networks present a narrow memory continuum (Fig.~\ref{fig:MESH_CAM_resutls}b,~\ref{fig:existing_nets}e), however they have a very low capacity. 

The bipartite expander Hopfield network~\cite{chaudhuri2019bipartite} can be used to perform robust label retrieval from noisy or partial pattern cues, for an exponentially large number of arbitrary patterns (Fig.~\ref{fig:existing_nets}b). However, the nature of memory explored in this network is familiarity or labeling, not reconstruction. Thus the information per pattern is very small, regardless of the number of stored patterns.

Dense (`Modern') Hopfield networks are recently proposed variants of the Hopfield model that involve higher-order interactions in place of the conventional pairwise interactions. These present a memory capacity that grows as $N^{K-1}$ or $\exp(N)$ dependent on the order of the interactions ~\cite{krotov2016dense,demircigil2017model,ramsauer2020hopfield}. Though a bipartite structure (Fig.~\ref{fig:schematics}b) with pairwise interactions can approximate higher-order interactions \cite{chaudhuri2019bipartite,krotov2020large}, the capacity of a CAM with such structure remains linear rather than exponential in the number of hidden nodes~\cite{krotov2020large}.  In fact, in~\cite{krotov2020large} the number of hidden units must {\em exactly} equal the number of memories, thus storage of a variable number of patterns requires a change of network architecture, rendering the network inflexible and unable to exhibit a memory continuum.

Overparameterized autoencoders can also act as a CAM, with patterns stored as the fixed points of iterations of the learned map of the autoencoder~\cite{radhakrishnan2020overparameterized} (Fig.~\ref{fig:schematics}c). A drawback of these CAMs is that autoencoders require extensive training through backpropagation, in contrast to the one-shot learning of associative memory models including all CAM models described above and MESH. Similar to other CAM models, overparametrized autoencoders also exhibit a memory cliff (Fig.~\ref{fig:existing_nets}f).
\section{MESH Exhibits a Near-Optimal CAM Continuum}
\label{sec:MESH_CAMC}

We present MESH, a memory architecture in which a single network, \emph{without reparametrization or restructuring}, can tradeoff in an online way increasingly many stored patterns for decreasing detail per pattern, thus populating the whole extent of the theoretical memory envelope,  Fig.~\ref{fig:schematics}g. 
MESH consists of two components, Fig.~\ref{fig:schematics}f: 1) a predefined ``memory scaffold'' --- implemented through a bipartite attractor network, with an $N_L$-dimensional label layer with $k$-hot activations, and an $N_H$-dimensional hidden layer --- which generates $\binom{N_L}{k}$ fixed points with large basins; and 2) a ``heteroassociative'' network --- in which $N_F$-dimensional inputs encoding up to $\binom{N_L}{k}$ arbitrary patterns in a feature layer are hooked onto the memory scaffold via hetroassociative learning. While we describe MESH in more detail in the following sections, here we first present its capabilities.
To probe memory recovery, the feature layer of MESH is cued with a corrupted version of a stored pattern. The retrieval dynamics (Fig.~\ref{fig:schematics}f) aims to return to the feature layer a cleaned up version of the cued pattern. MESH perfectly reconstructs upto $N_H$ arbitrary stored patterns $N_{patts}$ of size $N_F$ bits each when cued with clean or noisy patterns, Fig.~\ref{fig:MESH_CAM_resutls}a (see Fig.~\ref{fig:appendix_noisyOverlapMI} for noisy cues). 
When $N_{patts}$ is increased beyond this number, the network performs partial reconstruction of the stored patterns when cued by clean or noisy patterns, with a smooth decay in the quality of reconstructed patterns, Fig.~\ref{fig:MESH_CAM_resutls}a (Fig.~\ref{fig:appendix_noisyOverlapMI} for noisy cues). In this regime, every pattern is partially recalled, Fig.~\ref{fig:appendix_MIrelative}a (in contrast to models in which individual patterns are explicitly rewritten when adding new ones, thus any given pattern is either perfectly recalled or not at all, resulting in smooth degradation only on average~\cite{tyulmankov2021biological}. 


We next compare the information stored in MESH against CAM theoretical bounds \cite{gardner1988space, abu1989information} given by the total number of learnable synapses. In MESH, this bound is $MI_{total}^* =  {N_H(2N_F + N_L) }$. In practice $N_L\ll N_F$, and thus $MI_{total}^* \approx 2N_H N_F$. 

For patterns of length $N_F$, the CAM networks (Fig.~\ref{fig:existing_nets}) with a matched number of synapses may thus at best fully recall up to $2N_H$ patterns, but beyond exhibit a memory cliff. However, if a CAM network were to exhibit an optimal memory continuum, it should saturate the total information bound regardless of the number of stored patterns, with information per pattern per bit theoretically bounded by: 
\begin{align}
\label{Eqn:bound}
  MI_{perinbit}^* &= \left. MI_{total}^*\middle/ \left(N_{patts}\cdot \text{\# bits per pattern}\right)\right. \\ 
  &= \frac{N_H  (2N_F + N_L) }{N_{patts}N_F}\approx \frac{2 N_H}{ N_{patts}}.
\end{align}
Experimentally, we find that MESH nearly saturates this theoretical bound across a widely variable number of stored patterns, Fig.~\ref{fig:MESH_CAM_resutls}a (bound in dashed gray), in a single instance of the network without any architectural or hyperparameter changes. The per-input-bit mutual information matches the best-performing CAM models when the number of stored patterns is smaller than the traditional CAM capacity, and is dramatically bigger when the number of stored patterns is larger, Fig.~\ref{fig:MESH_CAM_resutls}b (also see Fig.~\ref{fig:appendix_MIrelative}b,c). The number of stored and partially retrievable patterns exceeds the traditional CAM pattern number capacity by orders of magnitude. Consistent with this result, the information per synapse at large pattern numbers in MESH is significantly larger than in existing CAM models,  Fig.~\ref{fig:MESH_CAM_resutls}c.

Asymptotically with an increasing number of stored patterns, the total information per synapse in MESH approaches a constant (Fig.~\ref{fig:MESH_CAM_resutls}c) --
demonstrating that the total information that can be successfully recovered by MESH in a network of fixed size is invariant to the number of stored patterns. This invariance dictates the smooth trade-off between MI per pattern (pattern richness) and number of patterns. 

We show in Sec.~\ref{sec:arbitrary_patts} that the number of feature layer bits, $N_F$, can scale as $\binom{N_L}{k} \gg N_L$. Further, we show (in Sec.~\ref{sec:memory_scaffold}) that the number of hidden layer bits, $N_H$, necessary to support MESH is constant for large $N_L$.  Thus, the number of learnable synapses in the network $N_H (2 N_F + N_L)$ scales as $\mathcal{O}(N_F)$. However, the total number nodes in MESH, $N_F + N_H + N_L$, also scales as $\mathcal{O}(N_F)$. Thus, in entirety, MESH is a network with $\mathcal{O}(N_F)$ nodes and synapses, indicative of a highly sparse network. Since the number of patterns perfectly reconstructed, $N_H$, is constant with respect to the number of synapses $\mathcal{O}(N_F)$, MESH spans a CAM continuum from storing $\mathcal{O}(1)$ patterns with $\mathcal{O}$(num of synapses) bits of information each, up to storing $\binom{N_L}{k}\sim\mathcal{O}(N_F)$ patterns with a nonzero level of detail (Fig.~\ref{fig:MESH_CAM_resutls}a).

In sum, MESH stores a constant amount of total information that is invariant to the number of stored patterns; this total information content is proportional to the theoretical synaptic upper bound of total information storage in CAMs and is distributed across patterns so that the information per pattern degrades gracefully with no memory cliff as a function of the number of stored patterns (see Fig.~\ref{fig:appendix_MIrelative}a for the feature recovery error distribution). Next, to understand the underlying mechanisms that permit this flexible memory performance, we will examine the functional properties of the two components of MESH: the memory scaffold in Section~\ref{sec:memory_scaffold} and the heteroassociative learning in Section~\ref{sec:arbitrary_patts}.


\section{Exponential Scaffold}
\label{sec:memory_scaffold}


\begin{figure*}[h]
\centering
\includegraphics[width=0.95\textwidth]{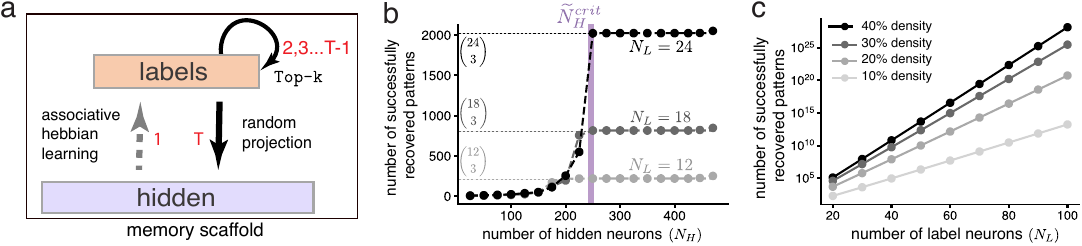}
\caption{\textbf{The memory scaffold part of MESH.} (a) The label-hidden layers form the scaffold, with discrete attractor states. (b) Capacity of the memory scaffold (successful recovery defined as $\leq$ 0.6\% error measured via the Hamming distance between the stored and recovered patterns, 
after 20\% input noise injected into the hidden layer). Different curves correspond to different label layer sizes for labels with a constant number of active bits ($k=3$). Given a critical number of hidden neurons, the memory scaffold achieves the maximum capacity given by $\binom{N_L}{k}$. All curves are averaged over 20 runs with different random projections ($W_{HL}$), error bars are too small to be visible. (c) Exponential capacity of the memory scaffold with $N_L$, assuming a constant density ($k/N_L$) of stored labels. 
}
\label{fig:mem_scaffold}
\end{figure*}


\begin{figure*}[h]
\centering
\includegraphics[width=0.95\textwidth]{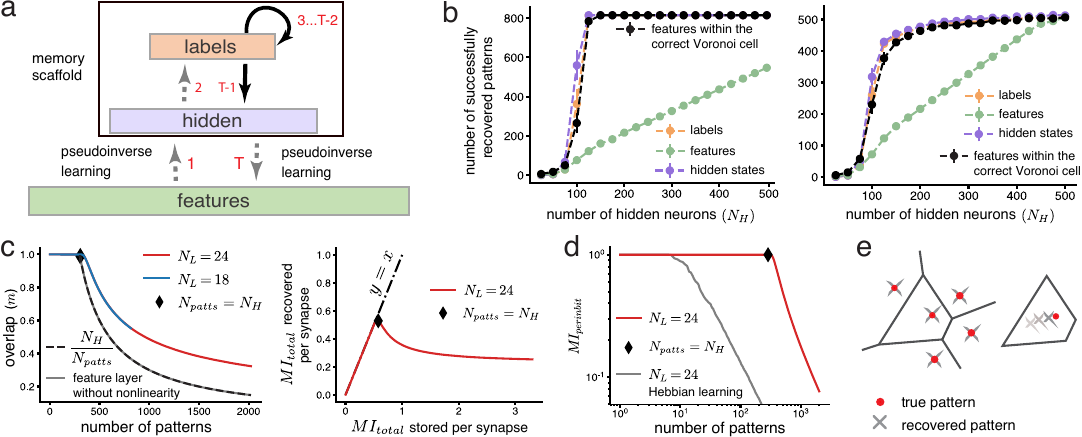}
\caption{\textbf{The heteroassociative part of MESH.} (a) Detailed architecture of MESH, with numbers in red indicating the order of dynamical updates. (b) Number of label, hidden and feature layer vectors perfectly recovered with zero error on cuing MESH with zero input noise (left) and  5\% input noise (right) in the feature states. Label and hidden layer vectors are always perfectly recovered, while feature layer vectors are perfectly recovered up to storage of $N_{H}$ patterns, with a continuum of recovered features beyond $N_{H}$ that continue to lie within the correct Voronoi cell. Here $N_L=18$, $k=3$ and $N_F = \binom{N_L}{k}$. (c) Left: Overlap between the stored and the recovered patterns. The corresponding curves for varying $N_L$ overlay with each other. Right: Total information per synapse (number of stored patterns increases along the x-axis) in MESH asymptotically approaches a constant. (d) Red curve: Mutual information (per input bit). Gray curve: Mutual information (per input bit) when $W_{HF}$ and $W_{FH}$ are trained using Hebbian learning, also producing a CAM continuum (although with a lower capacity) (e) Left: Voronoi cells for the stored patterns. When $N_{patts}\leq N_H$ the recovered pattern coincides with the stored pattern; Right: for $N_{patts}>N_H$ each pattern is partially recovered. However, the recovered pattern continues to lie in the correct Voronoi cell. All curves in (b), (c), (d) are averaged over 20 runs with different random projections ($W_{HL}$), error bars are too small to be visible.}
\label{fig:arbitrary_patterns}
\end{figure*}

The memory scaffold is a network that recurrently stabilizes a large number of prestructured states, exponential in the number of nodes. Further, these states have large basins of attraction, so enable denoising or clean-up of corrupted versions of these states. The predefined label layer states $l^\mu \in \{ 0,1\}^{N_L}$ are the set of $k$-hot patterns (each label state  is a vector with exactly $k$ bits set to ``1'', where $\mu$ is the pattern index), with $N_L$ the size of the small label (L) layer. 
The label layer projects with fixed dense random weights $W_{HL}$ to a much larger $N_H$-dimensional hidden (H) layer, Fig.~\ref{fig:mem_scaffold}a. These weights are drawn independently from a normal distribution with zero mean and unit variance ${W_{HL}}_{ij} \sim \mathcal{N}(0,1)$.
This defines the hidden-layer activations to be $h^{\mu} = \sgn (W_{HL} l^{\mu})$.
The return projections from the hidden (H) to the label (L) layer are learned through pairwise Hebbian learning between the set of predetermined label layer states, and these hidden-layer activations $h^{\mu}$ as given by Eq.~\ref{Eqn:return_proj}, where $C$ is a normalization term given by the number of predefined patterns, $\binom{N_L}{k}$. We assume that the label layer implements attractor dynamics through $k$-winners-take-all dynamics imposed by local recurrent inhibition \cite{majani1988k, rutishauser2011collective, wang2003k, yang1997dynamic}, enforcing through its dynamics that states remain $k$-hot at all times through a ``$\Top-k$'' nonlinearity (this $\Top-k$ nonlinearity can be replaced with a fixed threshold across all patterns; however the threshold would then have to be varied with $N_H$, Fig.~\ref{fig:appendix_Top-k}). 
\begin{equation}
\label{Eqn:return_proj}
    W_{LH} = \frac{1}{C}\sum_{\mu=1}^{C} l^{\mu} (h^{\mu})^T = \frac{1}{C}\sum_{\mu=1}^{C} l^{\mu} \sgn (W_{HL} l^{\mu})^T.
\end{equation}
Given a state $h(t)$, the memory scaffold states update as: 
\begin{align}
    l(t) &= \Top-k [W_{LH} h(t)], \label{eq:HtoL} \\
    h(t+1) &= \sgn [W_{HL} l(t)].\label{eq:LtoH}
\end{align}
The essential features that we desire for a memory scaffold are: i) high capacity --- the scaffold should have a large number of fixed points relative to the size of the network; ii) robust fixed points --- the basins of attraction for each of these fixed points must be sufficiently large to accommodate any perturbations induced while accessing the memory scaffold through the feature layer. As we show, each of the $\binom{N_L}{k}$ predefined states will form robust fixed points of the network with maximally large basins of attraction; and iii) strongly full rank --- the matrix formed by the first $N_{patts}$ scaffold states must be full rank for all $N_{patts}$. In our setup this almost always holds automatically for the (hidden) states (see Lemma~\ref{lem:Nhpatts} and Fig.~\ref{fig:ablation_rank}c). This feature is necessitated by properties of the heteroassociation that we address later in Theorem~\ref{thm:Nhpatts}.

\begin{theorem}\label{thm:scaffoldFP}
Given $N_L$ and $k$, there exists a critical number $N_H^{crit}(N_L,k)$ such that for $N_H > N_H^{crit}(N_L,k)$, all $\binom{N_L}{k}$ predefined $k$-hot label states are fixed points of the recurrent dynamics Eqs. (\ref{eq:HtoL},\ref{eq:LtoH}). Further, in the limit of $N_L\gg k\gg 1$, $N_H^{crit}(N_L,k)$ approaches $\widetilde{N}_H^{crit} = ck$ asymptotically, where $c$ is a constant that is solely determined by the largest permitted recovery error, and is independent of $N_L$ and $k$.
\end{theorem}
We prove this theorem under a simplifying assumption about the neural nonlinearity in Appendix~\ref{apx:scaffoldproof}. We also verify that, despite the simplification, the theoretical results agree qualitatively and quantitatively to numerical simulation in the full system (cf. Fig.~\ref{fig:appendix_memscaffold}b,c and Appendix~\ref{apx:scaffoldproof}).
We obtain directly as a corollary (proof in Appendix~\ref{sec:app_scaffold_converge})
\begin{corollary}\label{thm:onestep}
For $N_H>N_H^{crit}$, any vector $h(0)$ maps to a predefined scaffold state $h^\mu$ for some $\mu$ within a single iteration.
\end{corollary} 
As described in Sec.~\ref{sec:MESH_CAMC}, the hidden layer $H$ serves as an access point onto which the arbitrary patterns in the feature layer are hooked. Thus, we will primarily be interested in the robustness of these fixed points to perturbations to the hidden layer states $h^\mu$. 

\begin{theorem}\label{thm:basinsize}
For $N_H>N_H^{crit}$, all fixed points are stable, with equal-volume basins of attraction that are maximally large, i.e., the basin size is of the order of the size of the Voronoi cell of each pattern, $\text{Vol}[\{-1,1\}^{N_H}]/C$, where $C=\binom{N_L}{k}$ is the number of predefined scaffold states.
\end{theorem}

We prove this result in Appendix~\ref{sec:app_basinsize}. However, the presence of large volume basins of attraction alone does not guarantee robustness to perturbations. We show that these basins are convex in Appendix~\ref{apx:scaffoldconvex}, which then guarantees strong robustness to noise. We also note that the $\Top-k$ operation implemented as $k$-winners-take-all attractor dynamics can recurrently maintain the retrieved state over time. In this sense, the network is also able to hold a retrieved state as a short-term memory, as in Hopfield networks. 

Corresponding to Theorem~\ref{thm:scaffoldFP}, we experimentally observe that this bipartite memory scaffold can denoise states with high accuracy once the number of hidden neurons exceeds a critical value $N_{H}^{crit}(N_L,k)$. For a fixed value of $k$, this critical number varies weakly with the number of label neurons $N_L$ and approaches an $N_L$-independent constant $\widetilde{N}_H^{crit}$ in the limit of $N_L\gg k \gg 1$ (Fig.~\ref{fig:mem_scaffold}b, Appendix~\ref{apx:scaffoldproof}). Thus the critical number of hidden neurons can be considered to be independent of the number of stored patterns $C=\binom{N_L}{k}$ at constant $k$. One can therefore increase $N_L$ (while $N_L<N_H$) to obtain a capacity that at fixed $N_H$ grows rapidly with $N_L$ and $k$ as $\binom{N_L}{k}\sim (N_L)^k$. An even faster growth of large-basin scaffold states can be obtained by increasing the size of the label layer while holding the activity density ($d = k/N_L$) fixed (Fig.~\ref{fig:mem_scaffold}c). 
This results in a capacity that grows exponentially as $\binom{N_L}{k}\sim \exp(d N_L)$. Attaining this exponential growth in capacity requires an increase in $k$, which subsequently requires a  corresponding linear increase in $\widetilde{N}_H^{crit}$ (Fig.~\ref{fig:appendix_memscaffold}c, Appendix~\ref{apx:scaffoldproof}).

The number of large-basin stable states in this memory scaffold is far greater than the number of nodes and the number of synapses in this bipartite network, growing exponentially with the number of nodes. This does not violate CAM synaptic information bounds (since the stable states are predetermined rather than being arbitrary, and thus cannot transmit any information beyond the pattern index). We have thus demonstrated that the memory scaffold has a high capacity with large robust basins. The final requirement is for the hidden layer states to be strongly full rank:
\begin{lemma}\label{lem:Nhpatts}
The matrix $H$, constructed with $N_{patts}$ columns as the predefined hidden layer states $h^\mu$ is full rank for all $N_{patts}$, i.e., the matrix $H$ constructed over all $\binom{N_L}{k}$ patterns is strongly full rank. 
\end{lemma}
We argue that this result holds in Appendix~\ref{sec:app_hetero_htof}, and will be necessary for properties of heteroassociation (particularly Theorem~\ref{thm:Nhpatts}) to hold.

While we have provided a particular construction of a memory scaffold, other prestructured architectures with an exponentially high capacity of robust fixed points may also be used as memory scaffolds, such as Ref. \cite{chaudhuri2019bipartite}, or just the label layer itself with the recurrent $\Top-k$ nonlinearity (Fig.~\ref{fig:ablation}c, top). However, the scaffold states generated by these networks need to be full rank. Thereafter the states can always be reordered to be strongly full rank, although in our construction of the memory scaffold this holds automatically without any reordering.

We next demonstrate that the hidden layer can be used as an access point between arbitrary patterns and the memory scaffold, to hook external patterns onto the scaffold states.



\section{Heteroassociation of Arbitrary Patterns onto Scaffold}
\label{sec:arbitrary_patts}

\begin{figure*}[h]
\centering
\includegraphics[width=0.95\textwidth]{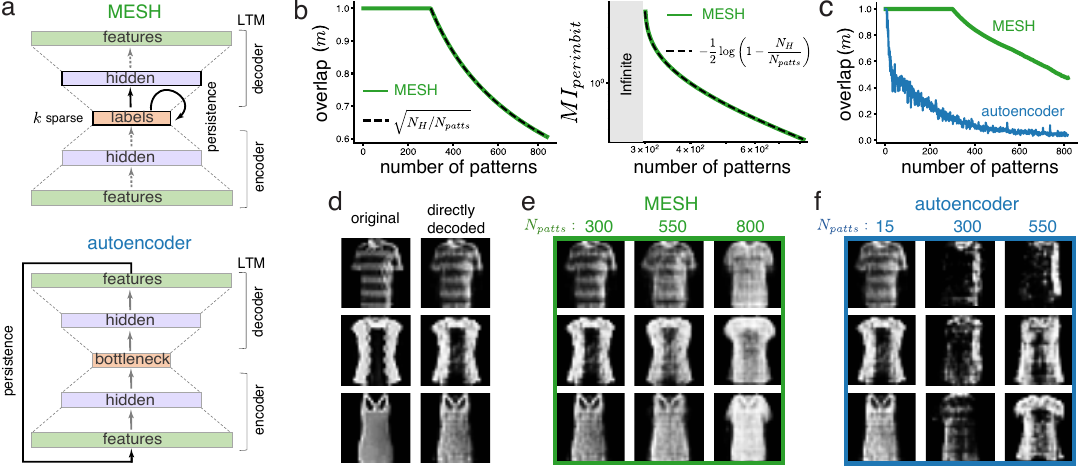}
\caption{\textbf{MESH exhibits a CAMC even when storing continuous valued patterns.} (a) MESH can be seen as a form of an autoencoder with constrained activations (boxes with black outlines), one-shot learned weights (grey-dashed arrows), and predetermined weights (black arrows); as opposed to autoencoders as CAMs trained with backprop. Despite lesser training, MESH enables equivalent performance below the memory cliff, and better performance beyond the memory cliff. (b) Overlap (left) and Mutual information (right) when MESH is trained on random continuous patterns. (c) Overlap when MESH and the overparameterized tail-biting autoencoder are trained on the Fashion MNIST dataset. While MESH shows a gradual degradation with increase in the number of stored patters, the autoencoder shows a catastrophic drop, shown visually in (e) and (f). (d) Left: original images from the Fashion MNIST dataset. Right: images directly decoded by a large decoder from the compressed feature representations. (e) Images decoded from the recovered feature representations stored in MESH trained on 300, 550, and 800 images respectively (left to right). MESH layer sizes: $N_L = 18$, $k = 3$, $N_H = 300$, $N_F = 500$. (f) Images decoded from the recovered feature representations stored in the overparameterized tail-biting autoencoder trained on 15, 300, and 550 images respectively (left to right). Layer sizes: 500, 300, 18, 300 and 500. All MESH curves in (b) and (c) are averaged over 20 runs with different random projections ($W_{HL}$), error bars are too small to be visible.}
\label{fig:applications}
\end{figure*}

The second component of MESH is bi-directional heteroassociative learning between the memory scaffold and inputs in the feature layer. The feature layer is the input and output of MESH: patterns to be stored are presented as random dense patterns\footnote{In analytical derivations, we consider dense random binary $\{-1,1\}$ patterns, although in practice this is not necessary (Sec.~\ref{sec:applications} shows examples of storage of continuous valued features).} which are then ``hooked'' onto one of the large-basin fixed points of the memory scaffold, Fig.~\ref{fig:arbitrary_patterns}a.
The heteroassociative weights are then set by the  pseudoinverse learning rule \cite{personnaz1985information},
\begin{align}
    W_{HF} &= H F^+  \;\;\; \text{and} \;\;\; W_{FH} = F H^+, \label{eq:WFH}
\end{align}
where the columns of $H$ and $F$ are the the predefined hidden layer states $h^\mu$ and input patterns $f^\mu$, respectively. Pseudoinverse learning can also be approximated through a (biologically plausible) online incremental learning mechanism \cite{tapson2013learning}, allowing weights to be learned as patterns are presented in an online or streaming setting. 
Note that the essential component of MESH is heteroassociative learning, not specifically the pseudoinverse rule. Heteroassociation through Hebbian learning, such that $W_{HF} = HF^T$ and $W_{FH} = FH^T$ also produces a CAM continuum in MESH, though as as seen in conventional Hopfield networks, pseudoinverse learning results in higher total stored information \cite{kanter1987associative, refregier1989improved, storkey1997increasing}, (Fig.~\ref{fig:arbitrary_patterns}d, gray curve). Furthermore, given a memory scaffold that perfectly recovers all hidden states, a single Hebbian heteroassociatve step is also sufficient for a continuum (see Appendix~\ref{sec:MI_hebb_theory}).

%

Presented with a noisy feature state $f(t)$ at time $t$, MESH dynamics are summarized as follows:
\begin{align}
    h(t) &= \sgn[W_{HF} f(t)], \\
    l(t) &= \Top-k[W_{LH} h(t)], \label{eq:memscaf1}\\
    h(t+1) &= \sgn[W_{HL} l(t)], \label{eq:memscaf2}\\
    f(t+1) &= \sgn[W_{FH} h(t+1)] \label{eq:frecon}.
\end{align}
Heteroassociative weights project noisy input patterns onto the hidden layer in the memory scaffold. The memory scaffold cleans up the received input by flowing to the nearest fixed point. This fixed point is decoded by the return projection to the feature layer, generating a non-noisy reconstruction of the input. To see why the heteroassociation with the memory scaffold allows for successful pattern storage and recovery, we examine the mapping from the feature layer to the memory scaffold, and then the recovery of the feature state from the scaffold. For the purpose of our arguments, we assume that the patterns being stored in the feature layer are random binary $\{-1,1\}$ patterns, and hence the matrix $F$ will be full rank. This allows the following results.
\begin{theorem}\label{thm:Hcorrect}
If the $N_F\times N_{patts}$ dimensional matrix $F$ is full rank, an input of clean feature vectors perfectly reconstructs the hidden layer states through heteroassociative pseudoinverse learning from the feature layer to the hidden layer, provided $N_{patts}\leq N_F$
\end{theorem}
Theorem~\ref{thm:Hcorrect} (proof in Appendix~\ref{sec:app_hetero_ftoh}) implies that cuing the network with unperturbed features stored in the memory results in perfect reconstruction of the predefined hidden layer states. Following the results in Sec.~\ref{sec:memory_scaffold}, if $N_H>N_H^{crit}$, reconstruction of the correct hidden states ensures that the correct predefined label states are also recovered (Fig.~\ref{fig:arbitrary_patterns}b, left).  The number of successfully recovered features is equal to the number of hidden neurons $N_H$, consistent with our description in Sec.~\ref{sec:MESH_CAMC}, a result that we now formalize. 

\begin{theorem}\label{thm:Nhpatts}
Assuming correctly reconstructed predefined hidden layer states that are strongly full rank, heteroassociative pseudoinverse learning results in perfect reconstruction of up to $N_H$ patterns in the feature layer.
\end{theorem}
This theorem (proof in Appendix~\ref{sec:app_hetero_htof}) also demonstrates the importance of the expansion of the label layer to a hidden layer of size $N_H$ --- setting up the predefined fixed points of the memory scaffold in a space that is higher dimensional than the label layer allows for perfect reconstruction of patterns up to the hidden layer dimensionality. This allows for the `knee' of the CAMC to be tuned as required by choosing an appropriate value of $N_H$ (Fig.~\ref{fig:appendix_MIvarylayers}a).


We now show that a CAM continuum exists for $N_{patts}>N_H$. We first show a result on the overlap between stored and recovered patterns before proving our main result on the mutual information of recovery of the CAM continuum.
\begin{theorem}\label{thm:ha_ov_continuum}
Assume that the memory scaffold has correctly reconstructed the predefined hidden layer states. Heteroassociative pseudoinverse learning from the hidden layer to the feature layer for $N_{patts}>N_H$ results in partial reconstruction of stored patterns such that the dot product between the stored patterns and the pre-sign-nonlinearity reconstruction of the stored patterns is $N_H/N_{patts}$ when averaged across all patterns.
\end{theorem}
Theorem~\ref{thm:ha_ov_continuum} (proof in Appendix~\ref{sec:app_overlapscaling}) along with Theorem~\ref{thm:Nhpatts} demonstrate the existence of the memory continuum --- for $N_{patts}\leq N_H$ the stored patterns are recovered perfectly, and for $N_{patts}>N_H$ the recovered patterns vary from the originally stored patterns in a smoothly varying fashion. However, Theorem~\ref{thm:ha_ov_continuum} only accounts for the overlap before the application of the sign nonlinearity; the sign nonlinearity in the feature layer only serves to additionally error correct the reconstructed patterns. This can be seen in Fig.~\ref{fig:arbitrary_patterns}c, left, where the gray curve presents the overlap before the application of the sign nonlinearity and is in close agreement to the theoretically expected result (dashed black curve). After this additional error correction, the mutual information recovered is then observed to asymptotically approach a $1/N_{patts}$ scaling as well, as seen in Fig.~\ref{fig:MESH_CAM_resutls}a. This can alternately be viewed as the mutual information per synapse approaching a constant as larger amounts of information are stored, Fig.~\ref{fig:arbitrary_patterns}c, right. Following Theorem~\ref{thm:ha_ov_continuum}, we note that the overlap between the true features and the recovered features is only a function of $N_H$ and $N_{patts}$. Thus, varying $N_L$ does not affect the magnitude of mutual information recovered and the corresponding curves for varying $N_L$ overlay with each other, Fig.~\ref{fig:arbitrary_patterns}c, left. 

Since a CAM continuum exists in MESH, when progressively more than $N_H$ patterns are stored in the network, the recovered pattern is progressively further from the true state. This is shown schematically in Fig.~\ref{fig:arbitrary_patterns}e where the Voronoi cells around each stored pattern are marked (i.e., the region closer to the stored pattern as compared to any other pattern): When the number of stored patterns is smaller than $N_H$ (Fig.~\ref{fig:arbitrary_patterns}e left), the recovered pattern corresponds exactly to the stored clean pattern. Storage of additional patterns up to the maximal $N_{patts} = \binom{N_L}{k}$, results in recovery of a different state that nevertheless remains within the Voronoi cell of the stored pattern (Fig.~\ref{fig:arbitrary_patterns}e right). 
In short, all recovered features remain within the correct Voronoi cell (corresponding to the uncorrupted stored pattern), which we verify numerically as the black curve in Fig.~\ref{fig:arbitrary_patterns}b, left. Thus, a CAM continuum in MESH results in approximate reconstruction of every pattern with gradual information decay for all patterns (Fig.~\ref{fig:appendix_MIrelative}a).



Until now, our theoretical results have only considered presentation of unperturbed versions of the stored patterns to the network. 
For up to $N_H$ patterns, presentation of corrupted versions of the stored features results in an approximate reconstruction of the hidden layer states, which through the memory scaffold dynamics then flows to the corresponding clean hidden and label state, Fig.~\ref{fig:arbitrary_patterns}b, right. This is then mapped back to the perfectly recovered stored pattern following Theorem~\ref{thm:Nhpatts}. 
For more than $N_H$ corrupted patterns, a similar process results in perfect recovery of the hidden layer and label layer states for a large number of patterns, although the capacity remains slightly smaller than the maximal number of patterns, $\binom{N_L}{k}$. 
\section{Continuous Patterns}
\label{sec:applications}


Next, we show that MESH exhibits a CAMC even when trained on continuous valued input patterns. To store continuous-valued patterns, the sign nonlinearity in Eq.~\ref{eq:frecon} is removed. To compare the stored and recovered patterns, we normalize them to unit $L_2$ norm before calculation of pattern overlap and recovered mutual information.


\textbf{Random Continuous Patterns}:
In this case we consider patterns $f^\mu$ such that for each $\mu$ and $i$, $f^\mu_i$ is independently sampled from a normal distribution with zero mean and unit variance. Since Lemma~\ref{thm:ha_ov_continuum} did not rely on any assumptions of pattern discreteness, the result extends to the case of continuous patterns. However, since we are normalizing the patterns before calculating the overlap, Eq.~\ref{eq:normoverlap} dictates the scaling of the overlap as $\sqrt{N_H/N_{patts}}$, consistent with the numerical results in Fig.~\ref{fig:applications}b, left. 
Since the stored patterns were drawn from a normal distribution, and assuming the recovered patterns are also distributed normally, the mutual information can be computed from the overlap ($m$) using the equation below (details in Appendix~\ref{sec:mi_cont_randn}), which is again in close agreement with the numerical results shown in Fig.~\ref{fig:applications}b (right), demonstrating the CAMC. Note that perfect reconstruction of continuous valued patterns (as in the case of $N_{patts}\leq N_H$) results in an infinite mutual information, $MI = -\log \left(1-m^2\right)/2 = -\log \left(1- N_H/N_{patts}\right)/2$.

\textbf{Fashion MNIST Dataset}:
To evaluate MESH on realistic images, we considered the toy problem of image storage from the Fashion MNIST dataset \cite{xiao2017fashion}. As the comparative model in this setting, we consider an equivalent tail-biting overparameterized autoencoder (Fig.~\ref{fig:applications}a, bottom) with the same number of nodes and synapses in each layer. MESH is an instance of this autoencoder with constrained label and hidden layer activations (Fig.~\ref{fig:applications}a, top), and weights trained through one-shot associative learning as opposed to backpropagation. Instead of the tail-biting connections, MESH has a $\Top-k$ recurrence on the label layer for the persistence of scaffold states. 

Since the images themselves have large pattern-pattern correlations, we found it beneficial for both MESH and the autoencoder to compress the dataset through a separate large autoencoder (details in Appendix~\ref{sec:fashion_mnist}). To visualize the memory storage, we pass the recovered patterns through the larger trained decoder to reconstruct the stored images of the shirts class (Fig.~\ref{fig:applications}d-f). Fig.~\ref{fig:applications}d shows a few samples of original images as well as images reconstructed directly from the compressed feature representations using the larger trained decoder. Fig.~\ref{fig:applications}e shows the images reconstructed through the decoder by using the feature representations recovered from MESH for varying numbers of stored patterns. When trained on $N_{patts}=N_H=300$ feature patterns, the image is reconstructed perfectly up to the quality of the larger decoder. As the number of stored feature patterns is increased ($N_{patts} > N_H$), the quality of image reconstruction gradually degrades. While the overparametrized autoencoder, Fig.~\ref{fig:applications}f, also accurately recovered the stored images when training on only 15 patterns, training on additional patterns results in the retrieved images becoming rapidly unrecognizable (as evident from Fig.~\ref{fig:applications}c,f).

\section{Discussion}



We have presented a CAM network, MESH, that exhibits a memory continuum (Fig.~\ref{fig:MESH_CAM_resutls}a), and can be used for storage/reconstruction, high capacity pattern labelling for recognition/familiarity detection, and locality sensitive hashing. While the convergence time of the Hopfield network scales as $\mathcal{O}(N_F^\gamma); \gamma \ll 1$ \cite{kohring1990convergence, frolov2000convergence}, MESH converges in a single step though an explicit $\Top-k$ operation, or within $\mathcal{O}(\log N_L)$ time when a $k$-winners-take-all dynamical attractor is used. Although here we have focused on one particular implementation of MESH, the necessary components are solely the memory scaffold and the heteroassociation which can be implemented in diverse ways provided the properties described in sections~\ref{sec:memory_scaffold} and~\ref{sec:arbitrary_patts} hold. To validate the necessity of each component in our implementation, we perform ablation studies in Appendix~\ref{sec:ablation_studies}. 



Several neural networks use a key-value mechanism to store and read memories from a lookup table \cite{graves2014neural, graves2016hybrid, sukhbaatar2015end, vaswani2017attention, le2019neural, banino2020memo}. MESH provides a neurally plausible architecture for the storage, lookup, and retrieval of memory states through a factorized key-value structure provided by the label and feature layers. Similarly, MESH also maps onto the entorhinal-hippocampal system in the brain, which factorizes sensory and spatial representations \cite{manns2006evolution, eichenbaum2014can}.

\clearpage
\nocite{langley00}
\bibliography{example_paper}
\bibliographystyle{icml2022}

\newpage
\appendix
\onecolumn
\section{Appendix}
\renewcommand\thefigure{\thesection.\arabic{figure}} 
\renewcommand\theHfigure{\thesection.\arabic{figure}} 
\setcounter{figure}{0}


This Appendix is structured as follows:
First, we present the quantification metrics and tools used to generate the numerical results presented in this paper in Apx.~\ref{apx:tools}. Then, in Apx.~\ref{apx:other_CAMS} we numerically characterize the performance of a wide range of other content addressable memory models, demonstrating their memory cliffs.
Thereafter, in Apx.~\ref{apx:scaffold_sec}-\ref{apx:heteroassoc_sec}, we provide theoretical guarantees of the results about MESH, first in Apx.~\ref{apx:scaffold_sec} focusing on the memory scaffold and in Apx.~\ref{apx:heteroassoc_sec} focusing on heteroassociative learning with the feature layer. In particular, in Apx.~\ref{apx:scaffold_sec} we prove that the setup of the memory scaffold described in Sec.~\ref{sec:memory_scaffold} results in a network with an exponentially large number of robust fixed points with large basins of attraction. Then, in Apx.~\ref{apx:heteroassoc_sec}, we first demonstrate that heteroassociative pseudoinverse learning will result in a memory continuum with the desired properties (as described in Sec.~\ref{sec:MESH_CAMC}), and then show that one may feasibly replace the pseudoinverse learning with simpler Hebbian learning and continue to obtain qualitatively similar results. Demonstrating the importance of each component of MESH, we consider a variety of ablation studies in Apx.~\ref{sec:ablation_studies} that establish the necessity of each compnent of MESH. Finally, we give additional details corresponding to applications of MESH for continuous valued patterns and its comparision with overparametrized autoencoders in Apx.~\ref{sec:fashion_mnist}

\begin{figure*}[h]
\centering
\includegraphics[width=\textwidth]{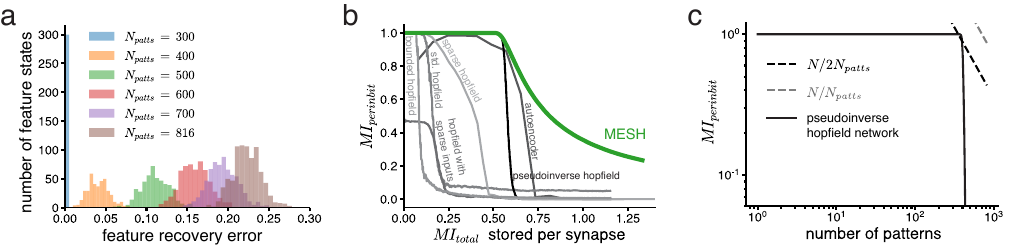}
\caption{(a) Feature recovery error forms a unimodal distribution, implying that each recovered pattern has a graceful degradation with increase in the number of stored patterns (every pattern still lies in the correct Voronoi cell). Here $N_L = 18$, $k = 3$, $N_H = 300$, $N_F = 816$. (b) Mutual information (per input bit) in existing networks relative to MESH as a function of total information stored per synapse. (c) Mutual information (per input bit) in a Pseudoinverse Hopfield network ($N = 708$). Gray dashed line shows the theoretical upperbound and black dashed line is the asymptotically proportional bound achieved by the network.}
\label{fig:appendix_MIrelative}
\end{figure*}

\section{Tools and Quantification Metrics}\label{apx:tools}
\subsection{Software and Data}

The source code for the models presented in this paper is made available at the following GitHub repository:\\
\url{https://github.com/FieteLab/MESH}

\subsection{Mutual Information}
\label{sec:mutual_info}


In this Appendix, unless otherwise specified, we use $\xi^\mu_i$ to represent the $i^\text{th}$ bit of the $\mu^\text{th}$ pattern stored in the network, and  $\sigma_i$ to represent the $i^\text{th}$ bit of the pattern recovered by the network. Here we restrict our analyses to the cases of random patterns such that bits of $\xi^\mu$ are independently sampled from i.i.d. random variables. This allows us to calculate information theoretic quantities for a single bit, and then scale the calculation by the pattern length to obtain the corresponding quantities for entire patterns. 

Further, for simplicity of notation, we overload $\sigma$ and $\xi$ to also represent the random variables from which the stored patterns and recovered patterns are being sampled.






We characterize the quality of pattern recovery by a network through the \emph{mutual information} between stored patterns $\xi$ and recovered patterns $\sigma$. For discrete random variables, the mutual information can be quantified as:

\begin{equation}\label{eqn:MI}
MI(\sigma; \xi) = H(\sigma) - H(\sigma|\xi),
\end{equation}
where $H(\sigma)$ is the information entropy of the recovered pattern $\sigma$, 
\begin{equation}\label{Eqn:marginal_entropy}
    H(\sigma) = - \sum_{\sigma} P(\sigma) \log P(\sigma)
\end{equation}
and $H(\sigma|\xi)$ is the conditional entropy of the recovered pattern given the stored pattern $\xi$,
\begin{equation}\label{Eqn:conditional_entropy}
    H(\sigma|\xi) = - \sum_{\xi} \sum_{\sigma} P(\sigma, \xi) \log P(\sigma|\xi).
\end{equation}

As we now show in the following sections, the mutual information can be explicitly computed for dense and sparse random binary patterns. 

\subsubsection{Dense binary patterns}
\label{sec:mi_binary}

For unbiased random binary \{-1,1\} patterns, 
\begin{equation*}
    H(\sigma) = - \frac{1}{2} \log \left(\frac{1}{2}\right) - \frac{1}{2} \log\left(\frac{1}{2}\right) = 1.
\end{equation*}
Further, since we assumed that each bit is independent, we obtain
\begin{equation}
    P(\sigma|\xi) = (1+m \sigma\xi)/2,
\end{equation}
where $m$ is the overlap between the stored and recovered pattern, $m = \frac{1}{N} \sum_{i} \sigma_i \xi_i$ \cite{dominguez2007information, bolle2000mutual}. Using Eq. (\ref{Eqn:conditional_entropy}), this can be used to obtain
\begin{align}
H(\sigma|\xi) &= - \frac{1}{2} \left( \frac{1+m}{2} \log \frac{1+m}{2} + \frac{1-m}{2} \log \frac{1-m}{2}\right) - \frac{1}{2} \left( \frac{1-m}{2} \log \frac{1-m}{2} + \frac{1+m}{2} \log \frac{1+m}{2}\right) \\
& = - \frac{1+m}{2} \log \left( \frac{1+m}{2}\right) - \frac{1-m}{2} \log \left( \frac{1-m}{2}\right). 
\end{align}    

Following Eq. (\ref{eqn:MI}) we thus obtain
\begin{equation}
MI(\sigma; \xi) = 1 + \frac{1+m}{2} \log \left( \frac{1+m}{2}\right) + \frac{1-m}{2} \log \left( \frac{1-m}{2}\right)     
\end{equation}

\subsubsection{Sparse binary patterns}
\label{sec:mi_sparse}

For sparse binary \{0,1\} patterns, let $p$ denote the fraction of ``1'' bits in the stored pattern (i.e., the average activity of the stored pattern). Let the average activity of the recovered pattern be denoted as $q = \sum_i\sigma_i/N$. 

Let $P_{1e}$ be the probability of error in a bit of $\sigma$ if the corresponding bit of $\xi$ is 1, and $P_{0e}$ be the error probability in a bit of $\sigma$ if the corresponding bit of $\xi$ is 0. Then,
\begin{align}
H(\sigma) &= - [ q\log(q) + (1-q)\log(1-q) ] \\
H(\sigma|\xi) &= - p [P_{1e} \log (P_{1e}) + 1-P_{1e} \log (1-P_{1e})] - (1-p) [P_{0e} \log (P_{0e}) + 1-P_{0e} \log (1-P_{0e})] \\
\end{align}

To obtain the probabilities $P_{1e}$ and $P_{0e}$, we compute the overlap $m$ and the average activity of the recovered pattern $q$ in terms of these probabilities as
\begin{align}
m &= (1/N)\sum_i \sigma_i \xi_i  = p(1-P_{1e}), \\
q &= \sum_i\sigma_i/N = p(1-P_{1e}) + (1-p)P_{0e} = m + (1-p)P_{0e}. 
\end{align}
These equations can then be solved to obtain
\begin{align}
P_{1e} &= 1 - m/p, \\
P_{0e} &= \frac{q-m}{1-p},
\end{align}

which can then be used to compute $MI(\sigma;\xi)$ using Eq. (\ref{eqn:MI}).

\subsubsection{Continuous random normal patterns}
\label{sec:mi_cont_randn}


The calculation of mutual information so far has been restricted to the case of discrete binarized patterns. For continuous valued patterns (as in Sec.~\ref{sec:applications}),entropy is ill-defined via Eq. (\ref{Eqn:marginal_entropy}). Instead, in this case we can defined the differential entropy as 
\begin{equation}
    h(X) = - \int_{-\infty}^{\infty} \phi(x)\log\phi(x)dx = \mathbb{E}[- \log\phi(x)],
\end{equation}
where $\phi(x)$ is the probability density function of the random variable $X$.

For random continuous patterns, as considered in Sec.~\ref{sec:applications}, the patterns are sampled from a normal distribution with zero mean and unit variance. This gives 
\begin{equation}
h(X) = \log \sqrt{2\pi e}.
\end{equation}  

The conditional entropy can similarly be calculated as
\begin{equation}
    h(X) = \log \sqrt{2\pi e (1-r^2)},
\end{equation}
where $r$ is the correlation coefficient between $X$ and $Y$. This can be used to obtain the mutual information
\begin{equation}\label{eqn:mi_cont_randn}
    MI(X;Y) = \log \sqrt {\frac{1}{1-r^2}}
\end{equation}




\subsection{Metrics}

We quantify the recovery error, i.e., the error between the stored pattern and the recovered pattern in the network, using Hamming distances. This recovery error is then used to quantitatively apply a recovery threshold to ascertain the capacity of the memory scaffold. 







After choosing a recovery threshold (see captions of Fig.~\ref{fig:mem_scaffold}b,c and Fig.~\ref{fig:arbitrary_patterns}b), the capacity of the network is defined as the largest number of patterns for which the average recovery error over all the stored patterns is below the chosen threshold.

\section{Performance of Existing CAM Models}\label{apx:other_CAMS}

\begin{figure*}[h]
\centering
\includegraphics[width=\textwidth]{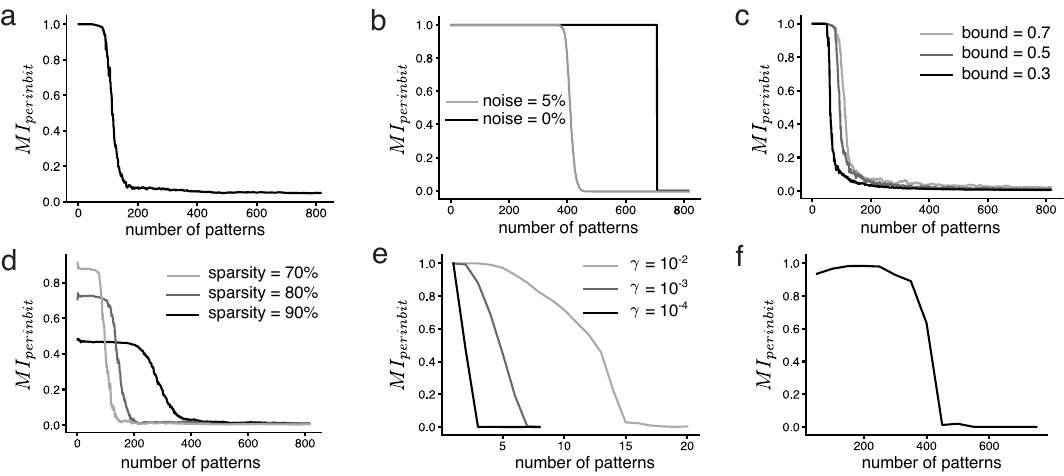}
\caption{ Mutual information (per input bit) between the stored and the recovered pattern as a function of number of patterns stored in the existing networks. (a) Hopfield network of size $N = 708$, synapses = $N^2$.(b) Pseudoinverse Hopfield network, tested with zero and non-zero input noise. Size of the network $N = 708$, synapses = $N^2$. 
(c) Hopfield network with bounded synapses trained with Hebbian learning on sequentially seen patterns. Size of the network $N = 708$, synapses = $N^2$. (d) Hopfield network with sparse inputs. Size of the network $N=708$, synapses = $N^2$, sparsity = $100(1-p)$. (e) Sparse Hopfield network. Size of the network $N$, synapse dilution $\gamma$, synapses = $\gamma \times N^2 = 10^5$ held constant for all curves. (f) Overparameterized Autoencoder (shown in Fig.~\ref{fig:schematics}c). Network layer sizes $N_{F} = 816$, $N_{H}=300$, $N_{B} = 18$. }
\label{fig:existing_nets}
\end{figure*}

To quantify and compare performance across models, in Fig.~\ref{fig:existing_nets}, we construct networks with $\approx5\times 10^5$ synapses, and measure the mean mutual information per-input-bit between the stored and the recovered patterns as a metric for the ability of the network to distinguish a memorized pattern from the recovered state (see Sec.~\ref{sec:mutual_info} for details). Unless otherwise specified, we average the MI across all patterns and normalize by the number of patterns and the number of bits per pattern (e.g., for storage of dense binary patterns, if all stored patterns are perfectly recovered, the MI will be unity, irrespective of the length of the patterns).

\section{Theoretical Results on the Memory Scaffold}\label{apx:scaffold_sec}

First, we prove that the the memory scaffold network has $\binom{N_L}{k}$ fixed points, while having only $\mathcal{O}(k N_L)$ synapses, establishing an exponentially large number of fixed points (D.1). Then, we demonstrate convergence to these fixed points occurs, in a single step (D.2), 
that each of these basins are maximally large (D.3), and finally demonstrate that these basins are convex (D.4), ensuring robustness of basins and protection against adversarial input.

\begin{figure*}[h]
\centering
\includegraphics[width=\textwidth]{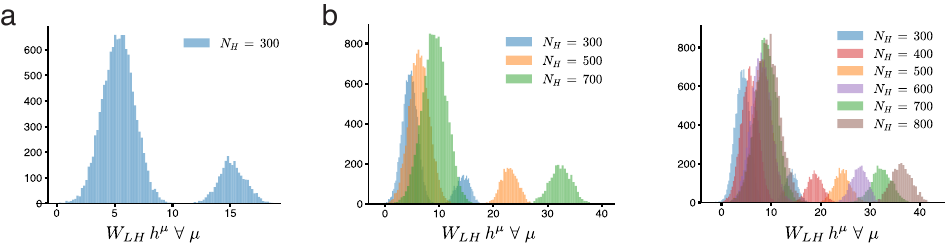}
\caption{Histograms of Label states before applying the $\Top-k$ nonlinearity in the dynamics of the MESH network. Here $N_L = 18$, $k = 3$, $N_F = 816$. (a) For a fixed number of hidden neurons  ($N_H = 300$) there's a clear threshold separating the $\Top-k$ states that can be implemented by a winner take all network. (b) Left: The choice of threshold can vary as $N_H$ varies, but a valid threshold always exists for any given $N_H$. Right: Threshold varies slowly with $N_H$ for large $N_H$; same threshold can be used for different $N_H$ since the $\Top-k$ states are increasingly distant and separable from the rest of the distribution.}
\label{fig:appendix_Top-k}
\end{figure*}

\begin{figure*}[h]
\centering
\includegraphics[width=\textwidth]{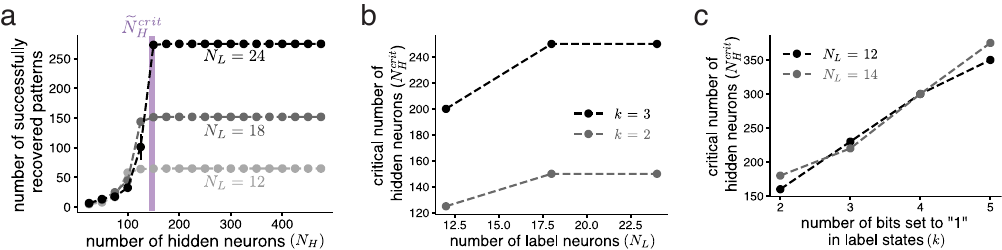}
\caption{(a) Capacity of the memory scaffold (computed with 20\% input noise injected in the hidden layer, and allowing up to a 0.6\% recovery error measured via the Hamming distance between the stored and recovered patterns). Different curves show the capacity corresponding to different sizes of the label layer for labels with a constant number of active bits ($k=2$). (b) Variation of the critical number of hidden neurons with respect to the number of label neurons. Asymptotically for large $N_L$, $N_H^{crit}$ becomes independent of $N_L$ (Sec.~\ref{apx:scaffoldproof}(c) Variation of the critical number of hidden neurons with respect to the number of active bits ($k$). $N_H^{crit}$ increases linearly with $k$.}
\label{fig:appendix_memscaffold}
\end{figure*}

\subsection{Heuristic Justification of Theorem~\ref{thm:scaffoldFP}} \label{apx:scaffoldproof}
We first provide broad qualitative justification for why the memory scaffold may be capable of storing such a large number of fixed points, before presenting a mathematical proof in a simplified setting. 

Unlike associative memory in the usual context of random patterns, note that the hidden layer states are determined by a random projection of the predefined label states. 
As a result the hidden layer states inherit similar pattern-pattern correlations as the predefined label layer states. This allows for Hebbian learning to act more efficiently in learning pairwise correlation resulting in a high capacity. Indeed, while in Hopfield networks any given fixed point is destabilized due to interference from other fixed points, the shared pattern-pattern correlations in the memory scaffold result in the interference terms being positively correlated with each fixed point (as can be seen in Fig.~\ref{fig:appendix_Top-k}).

To show this result more quantitatively, recall that
\begin{equation}\label{eq:full_dyn}
    l(t+1) = \Top-k[W_{LH}h(t)].
\end{equation}
Here, $\Top-k(x)$ is the nonlinear function that ensures that the vector $\Top-k(x)$ has exactly $k$ nonzero elements that are set to ``1'', and all remaining elements are set to ``0''. Further, the $k$ nonzero elements correspond to the same indices in the vector $x$ as the $k$ largest elements in $x$. Note that such a function can be constructed through recurrent connections in a neural circuit with a domain of all real-valued vectors\cite{majani1988k}.

Corresponding to the state $h^\mu$, consider the pattern $h(t) = h^\mu + \zeta$, where $\zeta$ represents a random noise vector. For simplicity, we assume that $\zeta$ is a continuous valued vector\footnote{this can be thought to correspond to discrete binary noise in the input feature vector, since the binary feature vector noise when mapped to the hidden layer will be continuous valued. This is not strictly true due to the sign nonlinearity in the hidden layer however.} whose each component is drawn independently from a normal distribution with zero mean and variance $\epsilon^2$. 

\begin{figure*}[h]
\centering
\includegraphics[width=0.45\textwidth]{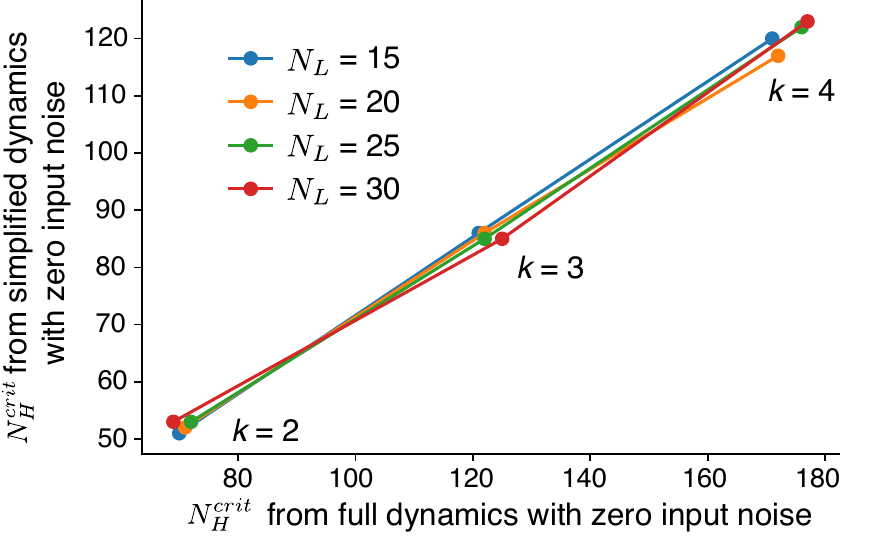}
\caption{$N_H^{crit}$ estimated numerically through the full MESH dynamics compared with the simplified system dynamics obtained by ignoring the sign nonlinearity in the hidden layer (cf. Eq. \ref{eq:simplified_dyn}). In both cases $N_H^{crit}$ with zero injected input noise, and by allowing an error corresponding to a hamming distance of $10^{-3} N_L$ when averaged across all scaffold states. See Sec.~\ref{apx:scaffoldproof} for more details.}
\label{fig:apx_approximation}
\end{figure*}

From $h(t)$, we aim to recover $l(t+1) = l^\mu$ via the mapping $W_{LH}$. For ease of notation, we denote the prespecified random projection $W_{HL}$ as $W$. Now, from the definition of $W_{LH}$ and $h^\mu$,
\begin{align}
    l(t+1) &= \Top-k[ LH^T (\sgn(W l^\mu)+\zeta) ], \\
           &= \Top-k[L\sgn(L^T W^T)\sgn(W l^\mu)/N_H + L\sgn(L^T W^T)\zeta/N_H],
\end{align}
where we add a scaling factor $1/N_H$ that leaves the $\Top-k$ calculation unchanged, but will be useful for normalization of random variables later in our calculation.
For analytic simplicity, we make the assumption that the sign nonlinearities in the above equation can be ignored. While this is a gross simplification, the obtained result is broadly consistent with the numerical observations in Sec.~\ref{sec:memory_scaffold}. This approximates the above equation to
\begin{equation}\label{eq:simplified_dyn}
    l(t+1) = \Top-k[ L L^T W^T W l^\mu/N_H + L L^T W^T\zeta/N_H] = \Top-k[A l^\mu + Z],
\end{equation}
where $A = (L L^T) (W^T W/N_H)$, and $Z =  L L^T W^T\zeta/N_H$. We numerically verify that this approximation is not unreasonable by estimating the minimum value of $N_H$ such that the hamming distance between $l(t+1)$ and $l(t) = l^\mu$ averaged over all $\mu$ is less than $10^{-3}\times N_L$ for $\zeta=0$. We observe a clear correlation between $N_H^{crit}$ estimated in this fashion using the simplified dynamics Eq. (\ref{eq:simplified_dyn}) versus using the full nonlinear dynamics Eq. (\ref{eq:full_dyn}) as shown in Fig.~\ref{fig:apx_approximation}, justifying the use of this approximation.

Since each element of the $N_H\times N_L$ matrix $W$ was drawn independently from a normal distribution with unit variance, $W^T W$ and hence $A$ can be treated as a matrix random variable. Further, as we show more precisely, the symmetry of the bit permutations across all patterns entails that each diagonal element of $A$ with i.i.d. and similarly each off-diagonal element of $A$ will be i.i.d. Let the i.i.d. variables on the diagonal be denoted as $\mathcal{X}_d^{i}$ for $i\in\{1\hdots N_L\}$, and the off-diagonal i.i.d. variables be denoted as $\mathcal{X}_f^{ij}$ for $i,j\in\{1\hdots N_L\}$, $i\neq j$.
\begin{equation*}
    A  = \begin{pmatrix} \mathcal{X}_d^{1} & \mathcal{X}_f^{12} & \hdots \\ \mathcal{X}_f^{21} & \mathcal{X}_d^2 & \hdots \\ \vdots & & \ddots \end{pmatrix}.
\end{equation*}
From the same bit-permutaion symmetry, we can also argue that the each component of the vector $Z$ will be sampled from i.i.d. random variables, which we denote as $\mathcal{Z}^i$.

Without loss of generality, let $l^\mu_i=1$ for $i\in\{1,\hdots,k\}$ and $l^\mu_i=0$ otherwise. Since a $\Top-k$ nonlinearity can be constructed to be valid over the domain of all real-valued inputs\cite{majani1988k}, it will thus suffice to require that the matrix $A$ acting on the vector $l^\mu$ when added to the noise vector $Z$ results in a vector with the value at the first $k$ indices to be larger than all other values. We first examine the constraints of the distributions $\mathcal{X}_d$, $\mathcal{X}_f$ and $\mathcal{Z}$ that allow for this to hold. In calculating $A l^\mu + Z$ we obtain a vector whose $i^\text{th}$ component is given by
\begin{equation}\label{eq:apx_firstk}
    \mathcal{X}_d^{i} + \sum_{1\leq j \leq k; j\neq i}\mathcal{X}_f^{ij} + \mathcal{Z}^i,
\end{equation}
for $1\leq i \leq k$, and 
\begin{equation}\label{eq:apx_afterk}
    \sum_{1\leq j \leq k}\mathcal{X}_f^{ij} + \mathcal{Z}^i,
\end{equation}
for $i>k$. Since we require that this vector has it's first $k$ elements larger than the remaining $N_L-k$ elements, we are interested in the probability 
\begin{equation}\label{eq:apx_Top-kcond}
    P(l(t+1)=l^\mu) = P(\text{Eq. (\ref{eq:apx_firstk}) - Eq. (\ref{eq:apx_afterk})} > 0) 
\end{equation}
Here we make an additional approximation for ease of analytic calculation: we will assume that the random variables $\mathcal{X}_d$, $\mathcal{X}_f$ and $\mathcal{Z}$ are normal variables (this approximation is valid in the large $N_H$ limit due to Central Limit Theorem). Let the mean and standard deviation of $\mathcal{X}_d$, $\mathcal{X}_f$ and $\mathcal{Z}$ be $\mu_d,\sigma_d$; $\mu_f,\sigma_f$; and, $\mu_Z,\sigma_Z$ respectively. In terms of these, the condition Eq. (\ref{eq:apx_Top-kcond}) can be rewritten as
\begin{equation}\label{eq:probTop-k}
    P(l(t+1)=l^\mu) =P(\mathcal{N}(\mu_d - \mu_f,\sigma_d^2 + (2k-1)\sigma_f^2 + 2\sigma_Z^2) > 0),
\end{equation}
with $\mathcal{N}(\mu,\sigma^2)$ representing a normal variable with mean $\mu$ and variance $\sigma^2$. This probability can be calculated as 
\begin{equation}\label{eq:prob_erf_Top-k}
    P(l(t+1)=l^\mu) =\frac{1}{2}\left[1+\erf\left(\frac{\mu_{AZ}}{\sigma_{AZ}\sqrt{2}}\right)\right]
\end{equation}
with $\mu_{AZ} = \mu_d - \mu_f$ and $\sigma_{AZ}^2 = \sigma_d^2 + (2k-1)\sigma_f^2+ 2\sigma_Z^2$. 

We now quantify more precisely the matrix random variable $A$, and will thereafter examine the random noise vector $Z$. From the definition of the set of patterns $l^\mu$, $L L^T$ can be shown to be a matrix with $\lambda_d = \binom{N_L-1}{k-1}$ on the diagonal, and $\lambda_f = \binom{N_L-2}{k-2}$ in each off-diagonal entry. 

Next, note that since each element of the $N_H\times N_L$ matrix $W$ was drawn independently from a normal distribution with unit variance, and thus $W^T W$ will have each diagonal element being distributed as the sum of the squares of $N_H$ standard normal variables, and each off-diagonal element will be distributed as the sum of the products of $N_H$ pairs of uncorrelated standard normal variables. Thus
\begin{equation}
    W^T W \sim \begin{pmatrix} \chi^2(N_H) & \mathcal{NP}(N_H) & \hdots \\ \mathcal{NP}(N_H) & \chi^2(N_H) & \hdots \\ \vdots & & \ddots\end{pmatrix},
\end{equation}
where $\chi^2(N)$ is the sum of $N$ i.i.d. $\chi^2$ distributions, and $\mathcal{NP}(N)$ is the sum of $N$ i.i.d. normal product distributions (i.e., the distribution of the product of two i.i.d. standard normal variables). Note that we have suppressed the indices on each matrix element, however it should be noted that each element is an independent sample from the distribution and are identical in distribution but not in value. 

In the large $N_H$ limit, each element of $W^T W$ is the sum of a large number of random variables and can hence be approximated as a normal distribution due to central limit theorem. Thus, $\chi^2(N_H)\sim\mathcal{N}(N_H,2 N_H)$, and $\mathcal{NP}(N_H)\sim\mathcal{N}(0,N_H)$. 

We thus treat $W^TW/N_H$ as a matrix random variable with elements on the diagonal being drawn from a distribution $\mathcal{D}$, having unit mean and a variance of $2/N_H$; and elements on the off-diagonal being drawn from a distribution $\mathcal{O}$, having zero mean and $1/N_H$ variance.

This can now be used to compute the distribution of the elements of $A$, in terms of the matrix elements of $LL^T$ and $W^T W$ to obtain
\begin{align}
    \mathcal{X}_d &= \lambda_d \mathcal{D} + \sum_{N_L-1\text{ terms}} \mathcal{O} \label{eq:Xd_dist}\\
    \mathcal{X}_f &= \lambda_f \mathcal{D} + \lambda_d \mathcal{O} + \sum_{N_L-2\text{ terms}} \mathcal{O}, \label{eq:Xf_dist}
\end{align}
where we again suppress indices over individual random variables but note that each random variable is i.i.d., including each summand term in the above expressions. 

This allows for $\mu_d,\sigma_d,\mu_f$ and $\sigma_f$ to be computed. 
\begin{eqnarray}
\mu_d &= \lambda_d, \\
\mu_f &= \lambda_f,
\end{eqnarray}
and,
\begin{eqnarray}
\sigma_d^2 = \left[2 \lambda_d^2 + \lambda_f^2 (N_L-1)\right]/N_H, \\
\sigma_f^2 = \left[2\lambda_f^2 + \lambda_d^2 + (N_L-2)\lambda_f^2\right]/N_H.
\end{eqnarray}
The quantity relevant to $\sigma_{AZ}$ is $\sigma_d^2 + (2k-1)\sigma_F^2$, which simplifies to
\begin{equation}
    \sigma_d^2 + (2k-1)\sigma_F^2 = \lambda_d^2 (2k+1) + \lambda_f^2(2k N_L -1)
\end{equation}

The variance $\sigma_Z^2$ can be computed in a similar fashion. First note that $W^T \zeta$ will be a random vector with each element constructed from the sum of $N_H$ i.i.d. normal product distributions multiplied by the scale of $\zeta$, i.e., $\epsilon$. Thus $W^T\zeta/N_H$ is identically distributed to $\epsilon \mathcal{O}$. Left multiplying this vector with $LL^T$ we obtain
\begin{equation}
    \mathcal{Z} = \epsilon\left[ \lambda_d  \mathcal{O} + \lambda_f  \sum_{N_L-1\text{ terms}}  \mathcal{O}\right],
\end{equation}
which gives
\begin{equation}
    \sigma_Z^2 = \epsilon^2\left[\lambda_d^2 + \lambda_f^2(N_L-1)\right]/N_H
\end{equation}

We can then use these to compute $\mu_{AZ}$ and $\sigma_{AZ}$ using $\mu_{AZ} = \mu_d - \mu_f$ and $\sigma_{AZ}^2 = \sigma_d^2 + (2k-1)\sigma_f^2 + 2\sigma_Z^2$. The ratio $\mu_{AZ}^2/\sigma_{AZ}^2$ simplifies to
\begin{equation}
    \frac{\mu_{AZ}^2}{\sigma_{AZ}^2} = \frac{N_H (N_L-k)^2}{(N_L-1)^2(2k+1+2\epsilon^2) + [2kN_L-1 + 2(N_L-1)\epsilon^2](k-1)^2}.
\end{equation}
This can then be directly inserted into Eq. (\ref{eq:prob_erf_Top-k}) to obtain the probability of reconstructing the correct label layer state.

Inverting the obtained expression allows for computation of $N_H^{crit}$,
\begin{equation}
    N_H^{crit} = \frac{c [(N_L-1)^2(2k+1+2\epsilon^2) + (2kN_L-1+ 2(N_L-1)\epsilon^2)(k-1)^2]}{(N_L-k)^2},
\end{equation}
where $c = 2 \left[ \erf^{-1}(1 - 2P) \right]^2$ and $P$ is the threshold selected for accuracy of the recovered pattern which increases with stricter thresholds. This allows us to estimate the critical number of hidden nodes (as a function of the number of label nodes, $N_L$, the number of ``on'' bits in the label, $k$ and the input noise $\epsilon$) beyond which the memory scaffold exists with all label states as fixed points.

For $N_L\gg k$ (as would be natural), the critical number of hidden nodes, $N_H^{crit}$ becomes independent of $N_L$ as it is asymptotically given by $c(2k+1+2\epsilon^2)$. This can be qualitatively seen in Fig.~\ref{fig:appendix_memscaffold}b. If we also assume that $\epsilon$ is smaller than $1$ and that $k\gg 1$ (as is also natural), then $N_H^{crit} \to \widetilde{N}_H^{crit} = \mathcal{O}(k)$. This has been verified qualitatively in numerical simulations in Fig.~\ref{fig:appendix_memscaffold}c, where $N_H^{crit}$ increases linearly with $k$, the number of on bits in the label layer. \footnote{Note that the analysis presented here assumes continuous valued noise in the hidden layer states, whereas the noise implemented in Fig.~\ref{fig:appendix_memscaffold} is the more relevant case of binary noise in the form of bit flips. As a consequence of this, the obtained values of $N_H^{crit}$ are hence not directly comparable.}.

These results thus demonstrate a crucial property of the Label layer-Hidden layer scaffold network --- it has $\mathcal{O}(N_L \widetilde{N}_H^{crit}) = \mathcal{O}(k N_L)$ synapses while having $\binom{N_L}{k}$ fixed points. Note that $\binom{N_L}{k}$ grows as $(N_L)^k$ with constant $k$, but crucially grows as $\exp(d N_L)$ if $k$ is proportional to $N_L$ with proportionality $d$. Thus, the number of fixed points grows exponentially faster than the number of synapses in the network, resulting in the network being useful as a memory scaffold for MESH.

\subsection{Memory Scaffold Dynamics Converge Within a Single Iteration}
\label{sec:app_scaffold_converge}

Here we provide the proof for the Corollary~\ref{thm:onestep} that claims that a single iteration convergence of the memory scaffold dynamics given $N_H > N_H^{crit}$.

\begin{proof}
This follows trivially from Eq. (\ref{eq:HtoL}) and Theorem~\ref{thm:scaffoldFP}: since the $\Top-k$ nonlinearity ensures that $l(0)$ will be a $k$-hot vector, and all $k$-hot vectors are predefined fixed points, thus $l(0)=l^\mu$ for some $\mu$. Correspondingly, in the next time step, $h(1) = \sgn[W_{HL}l^\mu] = h^\mu$, and the hidden layer state arrives at a fixed point within a single step starting from any vector. 
\end{proof}

\subsection{Memory Scaffold has Maximally Sized Basins of Attraction}
\label{sec:app_basinsize}

Theorem~\ref{thm:basinsize} suggests the existence of maximally sized basins of attraction that are equal in volume. Here we provide the proof for the same.

\begin{proof}
From Lemma~\ref{thm:onestep}, we see that the union of the basins about each of the predefined fixed points cover the entire space $\{-1,1\}^{N_H}$.
Note that each $h^\mu$ are equivalent, i.e., there is no special $\mu$ since each $l^\mu$ is equivalent up to a permutation of bits and $h^\mu$ are a random projection of $l^\mu$. Thus, $\{-1,1\}^{N_H}$ must be partitioned into basins with equal volume that are maximally large (and hence are of the same volume as the Voronoi cell about the fixed points).
\end{proof}

\subsection{Convexity of Scaffold Basins}\label{apx:scaffoldconvex}
Having shown the existence of maximally sized equi-volumed basins about each predefined memory scaffold state (Theorem~\ref{thm:basinsize}) is insufficient to guarantee robustness to noise. This is because large basins do not preclude the case of non-convex basins with basin boundaries arbitrarily close to the fixed points (which is the basis for adversarial inputs). Here we demonstrate that the obtained basins are convex, and thus the large basins must result in basin boundaries that are well separated from the fixed points themselves. 

Note that we are interested in the basins in the space $\{-1,1\}^{N_H}$, since the hidden layer is used as the access to the memory scaffold and noise robustness will hence be required there. The broad idea of the proof is as follows: first we demonstrate that perturbations in the binary hidden-layer space are equivalent to considering real-valued perturbations with small magnitudes in the label-layer space. Then we show that the $\Top-k$ nonlinearity on the labels results in convex basins in the label layer, which directly translates to convex basins in the hidden layer space. 

Consider a hidden layer state given by a small peturbation to a predefined hidden-layer fixed point $h^\mu$, which we donote as $h = h^\mu + \epsilon$. Since $h\in\{-1,1\}^{N_H}$, perturbations must take the form of bit-flips, and the vector epsilon must have values of either of $-2$, $2$ or $0$ at each component\footnote{More particularly, $\epsilon = -2 h^\mu \odot \hat \epsilon$, where $\odot$ represents pointwise multiplication, and $\hat \epsilon$ is a vector with a $1$ at the location of the bit flips and 0 otherwise, but the particular form of $\epsilon$ will not be essential to our argument}. Let $\delta$ denote the fraction of nonzero components of $\epsilon$. For small perturbations, $\delta \ll 1$. 
This is mapped to the label layer through $W_{LH}$ to obtain $\bar l$ before the application of the $\Top-k$ nonlinearity, where
\begin{align}
    \bar l &= W_{LH} h = W_{LH} [h^\mu + \epsilon] \\
           &= \bar l^\mu + W_{LH}\epsilon.
\end{align}
Note that $W_{LH}\epsilon$ will have a magnitude of approximately $\delta$ times the magnitude of $\bar l^\mu$, and further, the nonzero elements of $\epsilon$ are uncorrelated with $h^\mu$, and hence $W_{LH}\epsilon$ can be treated as a small real-valued perturbation to $\bar l^\mu = W_{LH}h^\mu$. 

If we can now show that the $\Top-k$ nonlinearity acting on a real-valued vector has a convex basin, that would indicate that all points near $l^\mu$ map to $l^\mu$, and since points near $h^\mu$ map to points near $l^\mu$, this would imply convexity of basins in $h$-space. 

Since all predefined label states are equivalent up to a permutation of indices, it will suffice to show that the basin about any fixed point is convex. Without loss of generality, we choose the fixed point $l^\mu$ whose first $k$ components are 1, and remaining components are 0. Let $p$ and $q$ be two real-valued vectors such that $\Top-k(p)=\Top-k(q) = l^\mu$, i.e., both $p$ and $q$ lie in the basin of attraction of $l^\mu$. Thus, $p_i>p_j$ and $q_i>q_j$ for all $1\leq i \leq k$ and $j>k$. Adding the two inequalities with coefficients $a$ and $(1-a)$, we obtain $a p_i + (1-a)q_i > a p_j + (1-a)q_j$ for all $1\leq i \leq k$ and $j>k$. Thus, $\Top-k(ap + (1-a)q) = l^\mu$. Thus for any two vectors $p$ and $q$ that lie in the basin of $l^\mu$, all vectors on the line from $p$ to $q$ also belong in the basin. Hence, the basin is convex in the label layer space, and as argued earlier this imposes convexity of basins in the hidden layer space.

Note that we have assumed in this proof that a $\Top-k$ function can be constructed with a domain over all real-valued vectors --- while in our implementation we explicitly constructed such a function, this can also be implemented in a neural network~\cite{majani1988k}.

\section{Theoretical Results on the Heteroassociative Learning}\label{apx:heteroassoc_sec}
Here we demonstrate that pseudoinverse learning first perfectly recovers the hidden layer states provided that $N_F>N_{patts}$ (in the noise-free case). Following the memory scaffold results proven earlier, reconstruction of the correct hidden layer states then results in correct retrieval of the corresponding label layer states. Next, we prove that for $N_{patts}<N_H$, the reconstructed feature layer states are also perfectly reconstructed, and for larger $N_{patts}$ the overlap of the stored and recovered patterns decays gracefully as described in Sec.~\ref{sec:MESH_CAMC}. We then prove that given an ideal memory scaffold, heteroassociative \emph{Hebbian} learning is also sufficient to obtain a memory scaffold with the same qualitative properties, with only a smaller prefactor on the memory capacity.

\subsection{Perfect Reconstruction of Hidden Layer States Through Heteroassociative Pseudoinverse Learning}
\label{sec:app_hetero_ftoh}

Here we provide the proof for Theorem~\ref{thm:Hcorrect}.

\begin{proof}
The projection of the stored patterns onto the hidden state is given by $W_{HF} F = H F^+ F = H P_F$,  where $P_F = F^+ F$ is an orthogonal projection operator onto the range of $F^T$. If $N_F \geq N_{patts}$, $F$ has linearly independent columns, and $P_F = \mathbf{1}$. Thus $W_{HF} F = H$. 
\end{proof}

\subsection{Perfect Reconstruction of $N_H$ Feature States Through Heteroassociative Pseudoinverse Learning}
\label{sec:app_hetero_htof}
To prove Theorem~\ref{thm:Nhpatts}, we first require lemma \ref{lem:Nhpatts}, \newline

While we do not prove this lemma, we note that $H = \sgn[W_{HL}L]$, and while $\rank(W_{HL}L) = \rank(L) = N_L < N_H$, the sign nonlinearity effectively acts as an independent random perturbation to the value of $H$ at each element, resulting in $H$ becoming full rank. This is numerically verified in Fig.~\ref{fig:ablation_rank}, where the rank can be seen to be min$(N_{patts}, N_H)$.
Theorem~\ref{thm:Nhpatts} can now be proved following a similar argument to Theorem~\ref{thm:Hcorrect}.

\begin{proof}
The projection of the hidden layer states onto the feature layer is given by $W_{FH} H = F H^+ H = F P_H$,  where $P_H = H^+ H$ is an orthogonal projection operator onto the range of $H^T$. For upto $N_{patts}\leq N_H$,  $P_H = \mathbf{1}$ and thus $W_{FH} H = F$. 
\end{proof}

As a numerical counter example to the importance of the full-rank property for all $N_{patts}$, consider the ablation study described in Fig.~\ref{fig:ablation}c where the label layer is considered as a memory scaffold in itself. Since a trivially chosen random ordering of the labels is not strongly full rank (cf. Fig.~\ref{fig:ablation_rank}c), the resulting capacity is significantly smaller than what might be expected for MESH. If we do order the label layer patterns in a specific sequence to satisfy the full-rank property for all $N_{patts}$, a higher, MESH-like capacity is recovered (see Fig.~\ref{fig:ablation_rank}b).

\subsection{Overlap in MESH Scales as $1/N_{patts}$}
\label{sec:app_overlapscaling}

 Theorem~\ref{thm:ha_ov_continuum} states that the overlap between the stored and recovered patterns (ignoring the sgn non linearity) in MESH scales as $1/N_{patts}$. Here we present the proof for the same. 

\begin{proof}
Let $\bar{f}^\mu$ be the reconstruction of pattern $f^\mu$ in the feature layer before the application of the sign nonlinearity, i.e., $\bar{f}^\mu = W_{FH} h^\mu$. Correspondingly, let $\bar F$ be the matrix constructed with $\bar f^\mu$ as its columns, i.e., $\bar F_{i\mu} = \bar f^\mu_i$. 
In this notation, we wish to prove that $f^\mu \cdot \bar f^\mu /|f^\mu|^2= N_H/N_{patts}$.

As earlier, $\bar F = F P_H$. Since $N_{patts}>N_H$, $\rank(H) = N_H$, and the projection operator $P_H$ is thus no longer an identity operator. Instead, $P_H$ projects on to the $N_H$-dimensional hyperplane $\mathcal{S}_H$ spanned by the rows of $H$. 
Notationally, let $\bar f_i$ be the vector corresponding to the $i^\text{th}$ row of $\bar F$, and similarly, let $f_i$ be the vector corresponding to the $i^\text{th}$ row of $F$.
In this notation, the vectors $\bar f_i$ (i.e., the rows of $\bar F$) are the vectors obtained by projecting $f_i$ (i.e., the rows of $F$) onto $\mathcal{S}_H$. 

By construction $f_i$ are $N_{patts}$-dimensional random vectors with no privileged direction. Thus, $|f_i|^2$, the squared magnitude along each dimension, will on average be equally divided across all dimensions. Hence, on average, the component of $f_i$ projected onto $\mathcal{S}_H$ (i.e., $\bar f_i$) will have a squared magnitude of $N_H |f_i|^2/N_{patts}$ and thus $|\bar f_i| = |f_i|\sqrt{N_H/N_{patts}}$. However, $|\bar f_i|$ is also the cosine of the angle between $f_i$ and the hyperplane $\mathcal{S}_H$, and hence averaged over $i$,
\begin{equation}\label{eq:normoverlap}
    f_i \cdot \bar f_i = |f_i||\bar f_i| \sqrt{N_H/N_{patts}}= |f_i|^2 N_H/N_{patts}.
\end{equation} 
Note that $\sum_i (f_i\cdot \bar f_i) = \sum_\mu (f^\mu \cdot \bar f^\mu)$, and $\sum_i |f_i|^2 = \sum_\mu |f^\mu|^2$. Thus the above equation can be rewritten as
\begin{equation}\label{eq:fmudotbarfmu}
    f^\mu \cdot \bar f^\mu = |f^\mu|^2 N_H/N_{patts}
\end{equation}
\end{proof}

\begin{figure*}[h]
\centering
\includegraphics[width=0.7\textwidth]{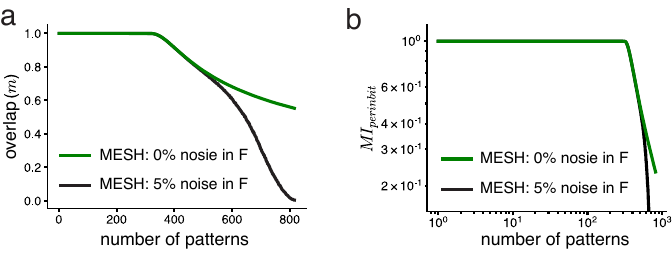}
\caption{(a) Overlap between the stored and the recovered patterns with 0\% and 5\% input noise in the feature layer. (b) Mutual information (per input bit) with 0\% and 5\% input noise in the feature layer. Both (a) and (b) correspond to results when storing random binary feature vectors.}
\label{fig:appendix_noisyOverlapMI}
\end{figure*}

\begin{figure*}[h]
\centering
\includegraphics[width=\textwidth]{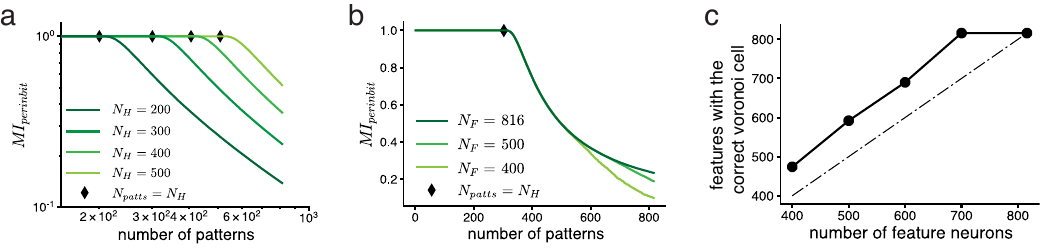}
\caption{(a) Effect of varying $N_H$ on Mutual Information (per input bit) in the MESH network. Here $N_L = 18$, $k = 3$, $N_F = 816$. (b) Effect of varying $N_F$ on Mutual Information (per input bit) in the MESH network. Here $N_L = 18$, $k = 3$, $N_H = 300$. (c) Number of recovered feature states in the correct Voronoi cell as a function of the number of feature neurons ($N_F$). Here $N_L = 18$, $k = 3$, $N_H = 300$, $N_{patts}=816$.}
\label{fig:appendix_MIvarylayers}
\end{figure*}

\subsection{One-Step Heteroassociation Leads to the Continuum Given a Perfect Memory Scaffold}
\label{sec:MI_hebb_theory}

\begin{figure*}[h]
\centering
\includegraphics[width=0.4\textwidth]{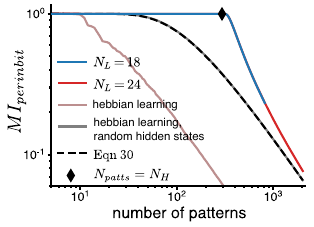}
\caption{Mutual Information (per input bit). Red and Blue curves: MESH memory continuum. Gray and Brown curves: weights $W_{FH}$ in MESH are trained using the Hebbian learning rule assuming perfect recovery of hidden states. Here gray curve corresponds to the case when hidden states are dense random binary.}
\label{fig:appendix_hebbian}
\end{figure*}

The memory continuum in MESH is a result of the one-step heteroassociation from the hidden to the feature layer, given a memory scaffold that perfectly recovers the hidden states. This holds irrespective of the nature of heteroassociation (pseudoinverse learning or hebbian learning). 

To illustrate this, we consider a simpler scenario where $W_{FH}$ is trained through Hebbian learning and the hidden layer states are assumed to be correctly reconstructed (corresponding to pseudoinverse learning from $F$ to $H$, Theorem~\ref{thm:Hcorrect}), making an additional approximation of the hidden layer states being dense random binary $\{-1,1\}$ patterns, we obtain an analytic expression for the mutual information per input bit as a function of the number of patterns (see Section~\ref{sec:Prob_error} for detailed proof) as follows 
\begin{equation}
\begin{aligned}
\label{Eqn:hebb_continuum}
MI_{perinbit} &= 1 + p \log p + (1-p) \log (1-p)
\end{aligned}
\end{equation}
for
\begin{equation}
\begin{aligned}
p = \frac{1}{2} \left[ 1 - \erf \left( \sqrt{\frac{N_H}{2N_{patts}}} \right) \right],
\end{aligned}
\end{equation}

where $\erf(x)$ is the Gauss error function. This result is in close agreement with numerical simulations (Fig.~\ref{fig:appendix_hebbian}, gray curve), and is asymptotically proportional to the MESH continuum. Furthermore, when the hidden states are random projections of label states (as in the original MESH network), rather than random dense states, the network still exhibits a continuum (Fig.~\ref{fig:appendix_hebbian}, brown curve). This shows that one step heteroassociation with Hebbian learning is itself sufficient for a memory continuum, given a perfect memory scaffold.

\subsubsection{MI for Hebbian learning}
\label{sec:Prob_error}
Given random binary dense hidden and feature states, if the weights from the hidden to the feature layer are trained using Hebbian learning, we get: 

\begin{equation}
\begin{aligned}
\label{Eqn:mi_hebb_1}
f_i (t_0 + \Delta t) &= \sgn\left[\frac{1}{N_H} \sum_{j=1}^{N_H} \sum_{\mu=1}^{N_{patts}} f_i^{\mu} h_j^{\mu} h_j^{\nu} \right] \\
&= \sgn \left[f_i^{\nu}  \left (\frac{1}{N_H} \sum_{j=1}^{N_H} h_j^{\nu} h_j^{\nu} \right) + \frac{1}{N_H} \sum_{\mu \neq \nu} \sum_{j} f_i^{\mu} h_j^{\mu} h_j^{\nu} \right] 
\end{aligned}    
\end{equation}

Here we have separated the pattern $\nu$ from all the other patterns. Next, we multiply the second term on the right-hand side by a factor $f_i^{\nu} f_i^{\nu} = 1$, and pull $f_i^{\nu}$ out of the argument of the sign-function since $f_i^{\nu} = \pm{1}$ : 

\begin{equation}
\begin{aligned}
\label{Eqn:mi_hebb_2}
f_i (t_0 + \Delta t) &= f_i^{\nu} \sgn \left[ 1 + \frac{1}{N_H} \sum_{\mu \neq \nu} \sum_{j} f_i^{\mu} f_i^{\nu} h_j^{\mu} h_j^{\nu} \right] = f_i^{\nu} \sgn[1 - a_{i\nu}] \\
where \;\; a_{i\nu} &= - \frac{1}{N_H}\sum_{\mu \neq \nu} \sum_{j} f_i^{\mu} f_i^{\nu} h_j^{\mu} h_j^{\nu}
\end{aligned}    
\end{equation}

If $a_{i\nu}$ is negative, the crosstalk term has the same sign as $f_i^{\nu}$ and does no harm. However, if it's positive and larger than 1 it changes the sign of $f_i^{\nu}$ and makes the bit $i$ of pattern $\nu$ unstable. Thus, the probability of error in recovering the true pattern $f_i^{\nu}$ is equal to the probability of finding a value $a_{i\nu} > 1$ for one of the neurons $i$.

Since both the hidden and feature state patterns are generated from independant binary random numbers $h_i^{\mu} = \pm{1}$ and $f_i^{\mu} = \pm{1}$ with zero mean, the product $f_i^{\mu} f_i^{\nu} h_j^{\mu} h_j^{\nu} = \pm{1}$ is also a binary random number with zero mean. The term $a_{i\nu}$ can be thought of as a random walk of $N_H (N_{patts}-1)$ steps and step size $1/N_H$ \cite{hertz2018introduction}. For a large number of steps, we can approximate the walking distance using a Gaussian distribution with zero mean and standard deviation given by $\sigma = \sqrt{(N_{patts}-1)/N_H} \approx \sqrt{N_{patts}/N_H}$ for $N_{patts} \gg 1$. The probability of error in the activity state of neuron $i$ is therefore given by:

\begin{equation}
\begin{aligned}
\label{Eqn:mi_hebb_3}
P_{error} &= \frac{1}{\sqrt{2\pi}\sigma} \int_1^{\infty} e^{\frac{-x^2}{2\sigma^2}} dx \approx \frac{1}{2} \left[ 1 - erf \left( \sqrt{\frac{N_H}{2N_{patts}}} \right) \right] \\
erf(x) &= \frac{2}{\sqrt{\pi}} \int_0^x e^{-y^2} dy 
\end{aligned}    
\end{equation}

Thus the probability of error increases with the ratio $N_{patts}/N_H$.

The mutual information between the stored and recovered feature states is thus given by:

\begin{equation}
\begin{aligned}
MI_{perinbit} &= 1 - \left(- [p \log p + (1-p) \log (1-p)] \right), \;\;\;  \texttt{(Using Eq.\ref{eqn:MI}}) \\
&= 1 + p \log p + (1-p) \log (1-p),
\end{aligned}
\end{equation}
where $p = P_{error}$.

\section{Ablation Studies}
\label{sec:ablation_studies}
\begin{figure*}[h]
\centering
\includegraphics[width=\textwidth]{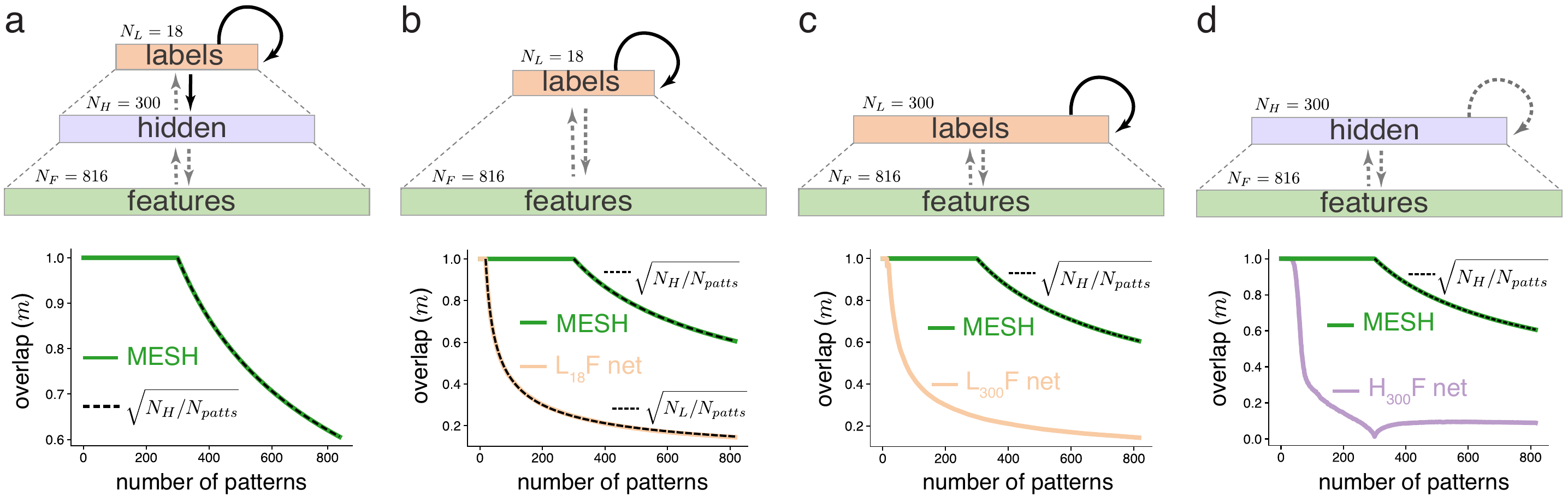}
\caption{Investigation of importance of each layer in MESH through ablation studies, examined for random continuous patterns. (a) MESH and its CAM continuum. (b) MESH without the hidden layer shows a much lower capacity and a $\sqrt{1/N_{patts}}$  rate of decay. (c) MESH without the hidden layer, where the label layer is scaled up to the size of the hidden layer (equal number of neurons and synapses as in a), still presents a low capacity along with a fast rate of decay since the memory scaffold formed by the label layer is not strongly full rank. See Fig.~\ref{fig:ablation_rank} for a similar ablation study where the labels have been reordered to be strongly full rank. (d) MESH without the label layer (with effectively same number of neurons and synapses as in a), with added recurrence in the hidden layer (trained similar to a Hopfield network) shows a catastrophic drop since the memory scaffold formed by the hidden layer doesn't satisfy the high capacity property. All curves in (a) - (d) are averaged over 20 runs, error bars are too small to be visible.}
\label{fig:ablation}
\end{figure*}

We investigate the significance of each layer in MESH through ablation studies (Fig.~\ref{fig:ablation}) that consider the removal of various components and examine the resultant change to the generated CAM continuum. 

\begin{figure}[h]
\centering
\includegraphics[width=0.52\textwidth]{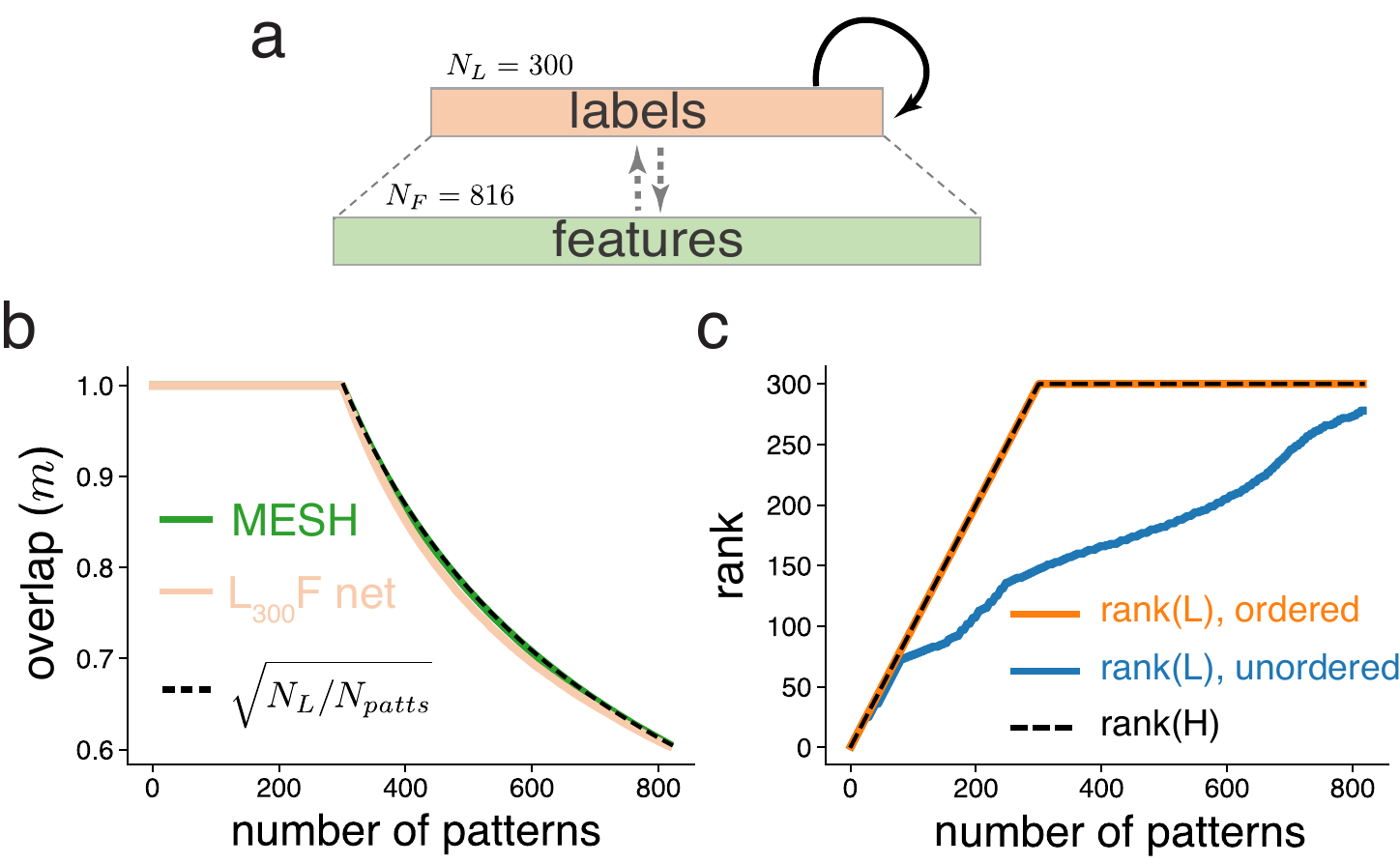}
\caption{(a) Ablated Mesh network without the hidden layer. (b) Performance of the ablated network when label layer patterns are ordered to be strongly full rank. MESH-like capacity is recovered. (c) Rank of label layer patterns when ordered and unordered. When ordered, the rank overlays that of the hidden layer patterns in MESH.}
\label{fig:ablation_rank}
\end{figure}

On ablating the hidden layer from MESH (Fig.~\ref{fig:ablation}b), the network shows a much lower capacity and much faster rate of decay in information. Naively, this is to be expected due to the significantly smaller rank of the memory scaffold states arising from lower scaffold state dimensionality.
However, on increasing the size of the label layer (to make it equivalent to the ablated hidden layer) (Fig.~\ref{fig:ablation}c), the network continues to demonstrate a diminished capacity, maintaining the rapid drop in information. 
This can be interpreted as occurring due to the absence of the strongly full rank property --- only special orderings of the label layer states $l^\mu$ result in a strongly full rank $L$ (Fig.~\ref{fig:ablation_rank}c, blue shows the rank of $L$ up to the first $N_{patts}$ columns for a randomly chosen ordering of $l^\mu$). Indeed, reordering the label layer states to be strongly full rank (corresponding to the orange curve in Fig.~\ref{fig:ablation_rank}c) restores a more MESH-like CAM continuum with a knee only at the dimensionality of the scaffold states, Fig.~\ref{fig:ablation_rank}b.

We establish the importance of a high capacity scaffold by considering an attempted memory scaffold generated by random hidden state activations along with recurrent Hebbian learning (this setup can alternatively be viewed as an ablation of the label layer with recurrence replacing the bipartite dynamics through the label layer). Similar to a Hopfield network, self-connections between nodes are set to zero by zeroing the diagonal of the recurrent connections matrix. As would be expected, this also does not lead to a memory continuum due to the ineffective memory scaffold, which is a Hopfield network that exhibits catastrophic forgetting at it's memory cliff.

\section{Compression Technique Used for Fashion MNIST}
\label{sec:fashion_mnist}

Since the Fashion MNIST images themselves have large pattern-pattern correlations, we found it beneficial for both MESH and the autoencoder to compress the dataset to extract lower-dimensional feature representations of the images through a separate large autoencoder. This large autoencoder was trained on all classes in the dataset except the ``shirts'' class. The encoder was then used to extract features of the ``shirts'' class which were used as the set of patterns to be stored in the MESH network and the overparameterized tail-biting autoencoder. Fig.~\ref{fig:applications}c shows the mean-subtracted overlap between recovered and stored patterns --- MESH continues to show a continuum, whereas the overparameterized autoencoder has a memory cliff at a very small number of patterns. 

Layer sizes of the larger autoencoder used to compress the dataset: 700, 600, 500, 600, 700. This autoencoder was used to compresse the $28 \times 28$ images to 500 dimensional features.

\end{document}